\documentclass{article}

\usepackage{ijcai16}
\usepackage{times}

\usepackage{graphicx}
\usepackage{booktabs}
\usepackage{multirow}
\usepackage{sidecap}
\usepackage{amsthm}
\usepackage{amsmath}
\usepackage{amsfonts}
\usepackage{url}
\usepackage{eqparbox}
\usepackage{arydshln} % for dashed line in tables 
\usepackage{bigstrut}
\usepackage[linewidth=1pt]{mdframed}
\usepackage{framed}
\usepackage{float}
\restylefloat{table}

\newif\ifarxiv

\arxivtrue

\newcommand{\namecite}[1]{\citeauthor{#1}~\shortcite{#1}}
\newcommand\citet{\namecite}

\usepackage{color}
\definecolor{purple}{rgb}{0.5,0,1}
\definecolor{dcyan}{rgb}{0.2,0.6,0.5}
\definecolor{darkgreen}{rgb}{0,0.25,0}

\newcommand{\addedA}[1]{{\color{black} #1}}
\newcommand{\addedD}[1]{{\color{black} #1}}
\newcommand{\addedT}[1]{{\color{black} #1}}

\theoremstyle{definition}
\newtheorem{definition}{Definition}

\newenvironment{myquote}{\list{}{\leftmargin=3ex\rightmargin=2ex}\item[]}{\endlist}

\allowdisplaybreaks
\newcommand{\tableilp}{TableILP}
\newcommand\lucene{IR}

\newcommand\salience{PMI}

\newcommand\praline{MLN}

\usepackage{array}
\newcolumntype{L}[1]{>{\raggedright\let\newline\\\arraybackslash\hspace{0pt}}m{#1}}
\newcolumntype{C}[1]{>{\centering\let\newline\\\arraybackslash\hspace{0pt}}m{#1}}
\newcolumntype{R}[1]{>{\raggedleft\let\newline\\\arraybackslash\hspace{0pt}}m{#1}}

\newcommand{\tableVar}{T_{i}}
\newcommand{\tableVarPrime}{T_{i'}}
\newcommand{\tableCell}{t_{ijk}}
\newcommand{\tableCellSameRow}{t_{ijk'}} % intra-table 
\newcommand{\tableCellPrime}{t_{ij'k'}} % intra-table 
\newcommand{\tableCellPrimePrime}{t_{i'j'k'}} % inter-table 
\newcommand{\header}{h_{ik}}
\newcommand{\option}{a_{m}}

\newcommand{\qCons}{q_\ell}
\newcommand{\qConsPrime}{q_{\ell'}}
\newcommand{\rowVar}{r_{ij}}
\newcommand{\rowVarPrime}{r_{ij'}}
\newcommand{\columnVar}{\ell_{ik}}
\newcommand{\columnVarPrime}{\ell_{ik'}}

\newcommand{\xOne}[1]{x\left(#1\right)}
\newcommand{\xTwo}[2]{y\left(#1, #2\right)}
\newcommand{\xFour}[4]{r\left(#1, #2, #3, #4\right)}

\newcommand{\setOf}[1]{\left\lbrace #1 \right\rbrace}

\title{Question Answering via Integer Programming over Semi-Structured Knowledge}

\newcommand\instA{$^\dagger$}
\newcommand\instB{$^\ddagger$}
\author{
  Daniel Khashabi\instA, Tushar Khot\instB, Ashish Sabharwal\instB, Peter Clark\instB, Oren Etzioni\instB, Dan Roth\instA\\
  \instA University of Illinois at Urbana-Champaign, IL, U.S.A.\\
  \textit{\{khashab2,danr\}@illinois.edu}\\
  \instB Allen Institute for Artificial Intelligence (AI2), Seattle, WA, U.S.A.\\
  \textit{\{tushark,ashishs,peterc,orene\}@allenai.org}
}

\begin{document}

\maketitle

%%%%%%%%%%%%%% ABSTRACT %%%%%%%%%%%%%%

\begin{abstract}
Answering science questions posed in natural language is an important AI challenge. Answering such questions often requires non-trivial inference and knowledge that goes beyond factoid retrieval. Yet, most systems for this task are based on relatively shallow Information Retrieval (IR) and statistical correlation techniques operating on large unstructured corpora. We propose a structured inference system for this task, formulated as an Integer Linear Program (ILP), that answers natural language questions using a semi-structured knowledge base derived from text, including questions requiring multi-step inference and a combination of multiple facts. On a dataset of real, unseen science questions, our system significantly outperforms (+14\%) the best previous attempt at structured reasoning for this task, which used Markov Logic Networks (MLNs). \addedA{It also improves upon a previous ILP formulation by 17.7\%.} When combined with unstructured inference methods, the ILP system significantly boosts overall performance (+10\%).  Finally, we show our approach is substantially more robust to \addedA{a simple answer} perturbation compared to statistical correlation methods.
\end{abstract}

\section{Introduction}
\label{sec:intro}

Answering questions posed in natural language is a fundamental AI task, with a large number of impressive QA systems built over the years. Today's Internet search engines, for instance, can successfully retrieve \emph{factoid} style answers to many natural language queries by efficiently searching the Web. Information Retrieval (IR) systems work under the assumption that answers to many questions of interest are often explicitly stated somewhere~\cite{Kwok2001ScalingQA}, and all one needs, in principle, is access to a sufficiently large corpus. Similarly, statistical correlation based methods, such as those using Pointwise Mutual Information or PMI~\cite{church1989}, work under the assumption that many questions can be answered by looking for words that tend to co-occur with the question words in a large corpus.

While both of these approaches help identify correct answers, they
%do not answer 
are not suitable for
questions requiring reasoning, such as chaining together multiple facts in order to arrive at a conclusion. Arguably, such reasoning is a cornerstone of human intelligence, and is a key ability evaluated by standardized science exams given to students.
% \emph{Can we build a natural language QA system that represents a step towards such intelligence?}
For example, consider a question from the NY Regents 4th Grade Science Test:
\begin{myquote}
{In New York State, the longest period of daylight occurs during which month? \ (A) June \ (B) March \ (C) December \ (D) September}
\end{myquote}

\begin{figure}[tb]
\centering
%\includegraphics[width=0.48\textwidth]{figures/ny-example}
%\begin{framed}
\includegraphics[trim=2.2cm 18.4cm 1cm 2.5001cm, clip=true, scale=0.66]{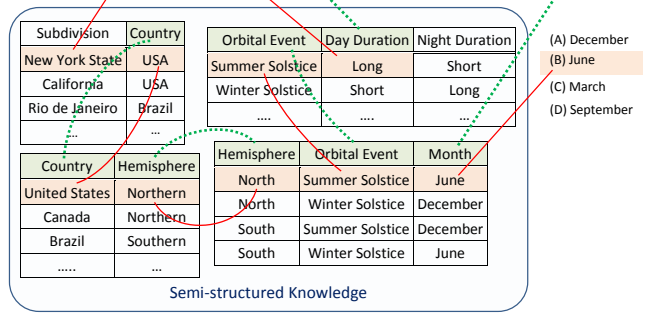}
%\end{framed}
%\vspace{-0.25in}
\caption{\tableilp\ searches for the best support graph (chains of reasoning)
%, shown with red and implicitly used green elements) 
connecting the question to an answer, in this case June.
%\ashish{seemed like too much detail at this point. simplified.}
%Red cells and edges denote active unary and pairwise elements in the support graph, respectively. Green nodes/edges are not part of the support graph, although they are used in its calculation. 
Constraints on the graph define what constitutes valid support and how to score it (Section~\ref{subsec:ilp}).} 
\label{fig:ny-example}
\end{figure}
\noindent
We would like a QA system that, even if the answer is not explicitly stated in a document, can \emph{combine basic scientific and geographic facts} to answer the question, e.g., New York is in the north hemisphere; the longest day occurs during the summer solstice; and the summer solstice in the north hemisphere occurs in June (hence the answer is June). Figure~\ref{fig:ny-example} illustrates how our system approaches this, with the highlighted support graph representing its line of reasoning.

% \begin{enumerate}
% \item Longest daylight corresponds to summer solstice.
% \item New York is in the United States.
% \item United States is in the northern hemisphere.
% \item Summer solstice in the northern hemisphere happens in the month of June.
% \end{enumerate}

Further, we would like the system to be \emph{robust under simple perturbations}, such as changing New York to New Zealand (in the southern hemisphere) or changing an incorrect answer option to an irrelevant word such as ``last'' that happens to have high co-occurrence with the question text.

To this end, we propose a structured reasoning system, called \tableilp,
%(QUestion-answering via Integer Programming),
%\ashish{better name?? TILP / TIP / QUIP / QAILP / QUAIL / ...}, 
that operates over a semi-structured knowledge base derived from text and answers questions by chaining multiple pieces of information and combining parallel evidence.\footnote{\addedA{A preliminary version of our ILP model was used in the ensemble solver of \namecite{aristo2016:combining}. We build upon this earlier ILP formulation, providing further details and incorporating additional syntactic and semantic constraints that improve the score by 17.7\%.}}
The knowledge base consists of \emph{tables}, each of which is a collection of instances of an $n$-ary relation defined over natural language phrases. E.g., as illustrated in Figure~\ref{fig:ny-example}, a simple table with schema \emph{(country, hemisphere)} might contain the instance \emph{(United States, Northern)} while a ternary table with schema \emph{(hemisphere, orbital event, month)} might contain \emph{(North, Summer Solstice, June)}. \tableilp\ treats lexical constituents of the question $Q$, as well as cells of potentially relevant tables $T$, as nodes in a large graph $\mathcal{G}_{Q,T}$, and attempts to find a subgraph $G$ of $\mathcal{G}_{Q,T}$ that ``best'' supports an answer option. The notion of best support is captured via a number of structural and semantic constraints and preferences, which are conveniently expressed in the Integer Linear Programming (ILP) formalism. We then use an off-the-shelf ILP optimization engine called SCIP~\cite{solver:scip} to determine the best supported answer for $Q$.

Following a recently proposed AI challenge~\cite{aristo2015:challenge}, \addedA{we evaluate \tableilp\ on unseen elementary-school science questions from standardized tests. Specifically, we consider a challenge set \cite{aristo2016:combining} consisting of all non-diagram multiple choice questions from 6 years of NY Regents 4th grade science exams.}
%, an expanded version of the dataset used by \namecite{aristo2015:mln}. 
In contrast to a state-of-the-art structured inference method~\cite{aristo2015:mln} for this task, which used Markov Logic Networks (MLNs)~\cite{richardson2006markov}, \tableilp\ achieves a significantly (+14\% absolute) higher test score. 
This suggests that a combination of a rich and fine-grained constraint language, namely ILP,
even with a publicly available solver
%with an industrial strength solver
is more effective in practice than various MLN formulations of the task. Further, while the scalability of the MLN formulations %explored by \namecite{aristo2015:mln} 
 was limited to very few (typically one or two) selected science rules at a time, our approach easily scales to hundreds of relevant scientific facts.
% \dan{This reads as if the only issue is computational power, but it really has to do with the formulation and with propsitionalization, right? Maybe worth a clarification sentence, either here or when discussing it in the eperimental section.} \ashish{agreed! will do}  
%
It also complements the kind of questions amenable to IR and PMI techniques, as is evidenced by the fact that a combination (trained using simple Logistic Regression \addedA{\cite{aristo2016:combining}}) of \tableilp\ with IR and PMI results in a significant (+10\% absolute) boost in the score compared to IR alone.

Our ablation study suggests that
%\daniel{Better adjective?: say ``chaining facts for reasoning'', ``reasoning on top of facts'', ``reasoning by joining facts'' or just ``reasoning'' instead of ``non-trivial reasoning''}
%\ashish{good suggestion!}
combining facts from multiple tables or multiple rows within a table plays an important role in \tableilp's performance. 
We also show that \tableilp\ benefits from the table structure, by comparing it with an IR system using the same knowledge (the table rows) but expressed as simple sentences; \tableilp\ scores significantly (+10\%) higher. Finally, we demonstrate that our approach is robust to a simple perturbation of incorrect answer options: while the simple perturbation results in a relative drop of 20\% and 33\% in the performance of IR and PMI methods, respectively, it affects \tableilp's performance by only 12\%.

%\peter{Can drop the paper overview I think ("The rest of the paper is structured as follows....")}\ashish{agreed! i personally don't like to include it... they can just flip pages :-)}
% The rest of the paper is structured as follows. After summarizing related work (Section~\ref{sec:related-work}), we describe our approach of treating QA as a subgraph selection problem (Section~\ref{subsec:qa-as-subgraph-selection}) and formulating this in the ILP framework (Section~\ref{subsec:ilp}). We then desribe results of an empirical evaluation (Section~\ref{sec:experiments}) and conclude with a summary (Section~\ref{sec:conclusion}).

%\input{3.related}
%%%%%%%%%%%%%% RELATED WORK %%%%%%%%%%%%%%

\section{Related Work}
\label{sec:related-work}

\addedA{\namecite{aristo2016:combining} proposed an ensemble approach for the science QA task, demonstrating the effectiveness of a combination of information retrieval, statistical association, rule-based reasoning, and an ILP solver operating on semi-structured knowledge. Our ILP system extends their model with additional constraints and preferences (e.g., semantic relation matching), substantially improving QA performance.}

A number of systems have been developed for answering factoid questions with short answers (e.g., ``What is the capital of France?'') using document collections or databases (e.g., Freebase~\cite{freebase}, NELL~\cite{nell}), for example~\cite{brill2002analysis,fader2014open,watson,ko2007probabilistic,yih2014semantic,yao2014information,zou2014natural}. However, many science questions have answers that are not explicitly stated in text, and instead require combining information together. Conversely, while there are AI systems for formal scientific reasoning (e.g., \cite{aura,isaac}), they require questions to be posed in logic or restricted English.
%rather than in natural language. 
Our goal here is a system that operates between these two extremes, able to combine information while still operating with natural language.

The task of Recognizing Textual Entailment (RTE)~\cite{dagan2010recognizing,dagan2013recognizing} is also closely related, as QA can be cast as entailment (Does {\it corpus} entail {\it question+answer}? \cite{Bentivogli2008TheSP}).
However, RTE has primarily focused on the task of {\it linguistic equivalence}, and has not addressed questions where some form of scientific reasoning is required. 
Recent work on Natural Logic~\cite{angeli2014NLI,maccartney2009:thesis} has extended RTE to account for the {\it logical structure} within language.
Our work can be seen as going one step further, to add a layer of structured reasoning on top of this; in fact, we use an RTE engine as a basic subroutine for comparing individual table cells in our ILP formulation.

ILP based discrete optimization has been successful in several NLP tasks~\cite{roth2004:ilp,chang2010discriminative,berant2010global,srikumar2011:srl,goldwasser2011:instructions}. While our ILP formulation also operates on natural language text, our focus is on the use of a specific semi-structured table representation for QA. \namecite{cohen2000:joins} studied tables with natural language text requiring soft matching, with a focus on efficiently computing the top few candidates given a database query.  In contrast, our system, given a natural language question, (implicitly) seeks to generate a query that produces the most supported answer.

% Most closely related to our work is the MLN-based system of \namecite{aristo2015:mln}, who proposed a number of different formulations of science QA in the MLN formalism~\cite{richardson2006markov} for probabilistic logic. They represented science knowledge as weighted first-order logic rules and formulated QA as the \#P-hard partial MAP problem~\cite{valiant1979complexity,roth1996hardness} of computing the marginal probability of an answer given the question and the science rules. Their system was severely limited by the well-known issue of grounding size explosion in MLNs, which meant it could work with very few facts at a time. In contrast, we formulate QA as an NP-equivalent discrete optimization problem, utilizing decades of advances in efficient LP engines.

% \ashish{Didn't get to include these:
% \cite{clark2014:akbc} \daniel{might be cite-able when introuce the tables?}
% \cite{yih2013question} \daniel{Doesn't matter; slightly irrelevant from out problem. }
% \cite{surdeanu2011learning}\daniel{Doesn't matter; not a fan of this work.}
% } \ashish{looks like OK on this front now}

%\input{4.ilp}
%%%%%%%%%%%%%% ILP APPROACH %%%%%%%%%%%%%%

\section{QA as Subgraph Optimization}

We begin with our knowledge representation formalism, followed by our treatment of QA as an optimal subgraph selection problem over such knowledge, and then briefly describe our ILP model for subgraph selection.
% ; full details of the ILP are deferred to the Appendix.

%%%%%%%%%%%%%%
\subsection{Semi-Structured Knowledge as Tables}
\label{subsec:tables}

We use semi-structured knowledge represented in the form of $n$-ary predicates over natural language text \addedA{\cite{aristo2016:combining}}. Formally, a $k$-column table in the knowledge base is a predicate $r(x_1, x_2, \ldots, x_k)$ over strings, where each string is a (typically short) natural language phrase. The column headers capture the table schema, akin to a relational database. Each row in the table corresponds to an instance of this predicate. For example, a simple country-hemisphere table represents the binary predicate $r_{\text{ctry-hems}}(c,h)$ with instances such as (Australia, Southern) and (Canada, Northern). Since table content is specified in natural language, the same entity is often represented differently in different tables, posing an additional inference challenge.

Although techniques for constructing this knowledge base are outside the scope of this paper, we briefly mention them. Tables were constructed using a mixture of manual and semi-automatic techniques. First, the table schemas were manually defined based on the syllabus, study guides, and training questions. Tables were then populated both manually and semi-automatically using \addedD{IKE \cite{ike2016}}, a table-building tool that performs interactive, bootstrapped relation extraction over a corpus of science text. In addition, to augment these tables with the broad knowledge present in study guides that doesn't always fit the manually defined table schemas, we ran an Open IE~\cite{Banko2007OpenIE} pattern-based \textit{subject-verb-object} (SVO) extractor from \namecite{clark2014:akbc} over several science texts to populate three-column Open IE tables. Methods for further automating table construction are under development.
\subsection{QA as a Search for Desirable Support Graphs}
\label{subsec:qa-as-subgraph-selection}

We treat question answering as the task of pairing the question with an answer such that this pair has the best support in the knowledge base, measured in terms of the strength of a ``support graph'' defined as follows.

Given a multiple choice question $Q$ and tables $T$, we can define a labeled undirected 
%\daniel{``undirected'' bugs me a bit; the directionality in some edges is important for us; however you might say that it comes into play when we assign weights, not in the graph itself.} \ashish{right, in our case I do see it more of a weight computation thing rather than inherent in the graph. hmm. will rethink.}
graph $\mathcal{G}_{Q,T}$ over nodes $\mathcal{V}$ and edges $\mathcal{E}$ as follows. We first split $Q$ into lexical constituents (e.g., non-stopword tokens, or chunks) $\mathbf{q} = \{\qCons\}$ and answer options $\mathbf{a} = \{\option\}$. For each table $\tableVar$, we consider its cells $\mathbf{t} = \{\tableCell\}$ 
%\daniel{Very minor issue; change \textbf{c} to \textbf{t} so that we don't confused it with ``question constituents''?}
%\ashish{done!}
as well as column headers $\mathbf{h} = \{\header\}$. The nodes of $\mathcal{G}_{Q,T}$ are then $\mathcal{V} = \mathbf{q} \cup \mathbf{a} \cup \mathbf{t} \cup \mathbf{h}$.
% where, with some abuse of notation, with 
For presentation purposes, we will equate a graph node with the lexical entity it represents (such as a table cell or a question constituent). The undirected edges of $\mathcal{G}_{Q,T}$ are $\mathcal{E} = ((\mathbf{q} \cup \mathbf{a}) \times (\mathbf{t} \cup \mathbf{h})) \cup (\mathbf{t} \times \mathbf{t}) \cup (\mathbf{h} \times \mathbf{h})$ excluding edges both whose endpoints are within a single table.

%\peter{I added this to try and explain *why* the entailment relation is introduced}
Informally, an edge denotes (soft) equality between a question or answer node and a table node, or between two table nodes. To account for lexical variability (e.g., that \emph{tool} and \emph{instrument} are essentially equivalent) and generalization (e.g., that a \emph{dog} is an \emph{animal}), we replace string equality with a phrase-level entailment or similarity function $w : \mathcal{E} \to [0,1]$ that labels each edge $e \in \mathcal{E}$ with an associated score $w(e)$. We use entailment scores (directional) from $\mathbf{q}$ to $\mathbf{t} \cup \mathbf{h}$ and from $\mathbf{t} \cup \mathbf{h}$ to $\mathbf{a}$, and similarity scores (symmetric) between two nodes in $\mathbf{t}$.\footnote{In our evaluations, $w$ for entailment is a simple WordNet-based \cite{miller1995wordnet} function that \addedA{computes the best word-to-word alignment between phrases, scores these alignments using WordNet's hypernym and synonym relations normalized using relevant word-sense frequency, and returns the weighted sum} of the scores. $w$ for similarity is the \addedA{maximum of the entailment score in both directions}. Alternative definitions for these functions may also be used.}
In the special case of column headers across two tables, the score is (manually) set to either 0 or 1, indicating whether this corresponds to a meaningful join. 
% In order to incorporate basic lexical information in this graph, such as a \emph{dog} is an \emph{animal} or that a \emph{tool} and an \emph{instrument} are essentially equivalent, we define a function $w : \mathcal{E} \to [0,1]$ that labels each edge $e \in \mathcal{E}$ with an associated entailment or similarity score $w(e)$. We use directional entailment scores from $\mathbf{q}$ to $\mathbf{t} \cup \mathbf{h}$ and from $\mathbf{t} \cup \mathbf{h}$ to $\mathbf{a}$, and a symmetric similarity score between two nodes in $\mathbf{t}$. The score between column headers in two tables is set to either 0 or 1, indicating whether this corresponds to a meaningful join. In our evaluations, $w$ is derived from Hypernym and Synonym relations in WordNet~\cite{miller1995wordnet}, although any other lexical resource may be used.

% \peter{I don't understand the intuition behind row-augmenting subgraph. Can you explain?}\ashish{added some explanation. helpful?} \peter{yes!}
Intuitively, we would like the support graph for an answer option to be connected, and to include nodes from the question, the answer option, and at least one table. Since each table row represents a coherent piece of information but cells within a row do not have any edges in $\mathcal{G}_{Q,T}$ (the same holds also for cells and the corresponding column headers), we use the notion of an augmented subgraph to capture the underlying table structure. Let $G = (V,E)$ be a subgraph of $\mathcal{G}_{Q,T}$. The \emph{augmented subgraph} $G^+$ is formed by adding to $G$ edges $(v_1,v_2)$ such that $v_1$ and $v_2$ are in $V$ and they correspond to either the same row (possibly the header row) of a table in $T$ or to a cell and the corresponding column header.
%\daniel{I didn't quite catch why we defined augmented graph (and why it is augmented only with row variables. )}

\begin{definition}
\label{def:support-graph}
A \emph{support graph} $G = G(Q,T,\option)$ for a question $Q$, tables $T$, and an answer option $\option$ is a subgraph $(V,E)$ of $\mathcal{G}_{Q,T}$ with the following basic properties:
%(1) $V \cap \mathbf{a} = \option$, (2) $V \cap \mathbf{q} \neq \phi$, (3) $V \cap \mathbf{t} \neq \phi$, (4) $w(e) > 0$ for all $e \in E$, (5) if $e \in E \cap (\mathbf{t} \times \mathbf{t})$ then there exists a corresponding $e' \in E \cap (\mathbf{h} \times \mathbf{h})$ involving the same columns, and (6) the row-augmented subgraph $G_r$ is connected.
\begin{enumerate}
\item $V \cap \mathbf{a} =  \lbrace \option \rbrace, \ V \cap \mathbf{q} \neq \phi, \ V \cap \mathbf{t} \neq \phi$;
\item $w(e) > 0$ for all $e \in E$;
\item if $e \in E \cap (\mathbf{t} \times \mathbf{t})$ then there exists a corresponding $e' \in E \cap (\mathbf{h} \times \mathbf{h})$ involving the same columns; and
\item the augmented subgraph $G^+$ is connected.
\end{enumerate}
\end{definition}

%\daniel{Having these bullets in the definition bug me tad bit. Here we define the ``support graph'' and later we talk about  `` the desirable support graph''. One can ask what's the point of defining ``support graph'', while we only care about the ``desirable support graph''? I suggest we keep the definition of the support graph a little informal, say by mentioning ``it shows the trace of reasoning using different elements in the problem and external knowledge''. And later we can say ``In the ILP formulation we intent to formulate and find a sensible support graph for a given question'' (hence we never talk about ``desirable support graph". )}\dan{I think it's ok to have a definition for support graphs; only say immediately after something like: A given question and a collection of tables give rise to a large number of support graphs, and the role of the inference process will be to choose the ``best" one under a definition of best developed next.}\ashish{I am with Dan on this; will add his phrasing}

A support graph thus connects the question constituents to a unique answer option through table cells and (optionally) table headers corresponding to the aligned cells. A given question and tables give rise to a large number of possible support graphs, and the role of the inference process will be to choose the ``best" one under a notion of \emph{desirable} support graphs developed next. We do this through a number of additional structural and semantic properties; the more properties the support graph satisfies, the more desirable it is.
%Our ILP formulation in the next section will make the measure of desirability or \emph{strength} of a support graph more precise.

%\daniel{I am worried that some readers might think that the whole ILP formulation is supposed to model the 4 properties mentioned earlier. However we know that the full formulation is very detailed and not easy to express in a few bullets. Maybe we can somehow say that the result of our ILP is an instance that belongs to the family of valid support graphs? In other words the space of feasible solutions for ILP program is much smaller that the space of the space of valid support graphs.}

% \tushar{Would be nice to somehow connect the proof graphs with multiple row inference, relation matching to fit the ablation study.}
% \daniel{I think these properties will/should be mentioned in the next section where we talk about ``desirable properties'' of the support graph. }\tushar{Agreed. At some point, this section was going to introduce some broad proof/support graph properties.} \ashish{Right, might revisit that tomorrow, but currently easier to leave it all in 3.3. But we should add some sentence about Relation Matching (as it appears in ablation) in 3.3. May be a bullet?}

%%%%%%%%%%%%%%
\subsection{ILP Formulation}
\label{subsec:ilp}

We model the above support graph search for QA as an ILP optimization problem, i.e., as maximizing a linear objective function over \addedA{a finite set of}
%integer- and continuous-valued \daniel{do we want to keep it? in the default setting no continuous variables are active} 
variables, subject to a set of linear inequality constraints. A summary of the model is given below.	
\ifarxiv 
	\footnote{Details of the ILP model may be found in the appendix.}
\else
	\footnote{Details of the ILP model \addedA{are omitted and may be found in the extended version of this paper \cite{tableilp2016:arxiv}.}}
\fi 
We note that  the ILP objective and constraints aren't tied to the particular domain of evaluation; they represent general properties that capture what constitutes a well supported answer for a given question.

\begin{table}[tb]
\centering
\setlength\tabcolsep{10pt}
\setlength\doublerulesep{\arrayrulewidth}
\small
%\begin{tabular}{|c|L{40ex}|} 
\begin{tabular}{c|l}
\textsc{Element} & \multicolumn{1}{c}{\textsc{Description}} \\ 
\hline\hline
\bigstrut[t] $\tableVar$ & table $i$ \\ 
%$\tableLen$ & Number of tables \\ 
$\header $  & header of the $k$-th column of $i$-th table \\
$\tableCell $ & cell in row $j$ and column $k$ of $i$-th table \\ 
$\rowVar$ & row $j$ of $i$-th table \\ 
$\columnVar$ & column $k$ of $i$-th table \\ 
$\qCons$ & $\ell$-th lexical constituent of the question $Q$ \\
%$\qLen$ & Length of a question \\ 
$\option$ & $m$-th answer option \\ 
\hline
\end{tabular}
\caption{Notation for the ILP formulation. }
\label{table:summary-notation}
\end{table}

Table~\ref{table:summary-notation} summarizes the notation for various elements of the problem, such as $\tableCell$ for cell $(j,k)$ of table $i$. All core variables in the ILP model are binary, i.e., have domain $\{0,1\}$. For each element, the model has a unary
%\daniel{drop ``activity'' in this sentence? And right before this sentence add ``We can a variable ``active'' if its value in our optimization problem is 1.''?}
%\ashish{dropped ``activity''; seems better to relate the variable's meaning to the support graph rather than to its value in the ILP}
variable capturing whether this element is part of the support graph $G$, i.e., it is ``active''. For instance, row $r_{ij}$ is active if at least one cell in row $j$ of table $i$ is in $G$. The model also has pairwise ``alignment'' variables, capturing edges of $\mathcal{G}_{Q,T}$. The alignment variable for an edge $e$ in $\mathcal{G}_{Q,T}$ is associated with the corresponding weight $w(e)$, and captures whether $e$ is included in $G$. To improve efficiency, we create a pairwise variable for $e$ only if $w(e)$ is larger than a certain threshold. These unary and pairwise variables are then used to define various types of constraints and preferences, as discussed next.

\addedD{ 
To make the definitions clearer, we introduce all basic variables used in our optimization in Table~\ref{table:ilp-variables}, and will use them later to define constraints explicitly. 
%Table~\ref{table:variables-and-description} summarizes our notation to refer to various elements of the problem, such as $\tableCell$ for cell $(j,k)$ of table $i$, as defined in Section 3. 
%As mentioned earlier, all our variables have domain $\{0,1\}$.
We use the notation $\xOne{.}$ to refer to a unary variable parameterized by a single element of the optimization, and $\xTwo{.}{.}$ to refer to a pairwise variable parameterized by a pair of elements. 
Unary variables represent the presence of a specific element as a node in the support graph $G$. For example $\xOne{\tableVar} = 1$ if and only if the table $\tableVar$ is active in $G$. Similarly, $\xTwo{\tableCell}{\qCons}= 1$ if and only if the corresponding edge is present in $G$, which we alternatively refer to as an \emph{alignment} between cell $(j,k)$ of table $i$ and the $\ell$-th constituent of the question. 

% \begin{table}[htb]
% \centering
% \small
% \begin{tabular}{|c|L{40ex}|} 
% \hline
%  \bigstrut[t] \textsc{Reference} & \bigstrut[t] \textsc{Description} \\ 
% \hline
% \bigstrut[t] $i$ & index over tables \\ 
% $j$ & index over table rows \\ 
% $k$ & index over table columns \\ 
% $l$ & index over lexical constituents of question \\
% $m$ & index over answer options \\
%  \hline
% \bigstrut[t] $\tableVar$ & table $i$ \\ 
% %$\tableLen$ & Number of tables \\ 
% $\header $  & header of the $k$-th column of $i$-th table \\
% $\tableCell $ & cell in row $j$ and column $k$ of $i$-th table \\ 
% $\rowVar$ & row $j$ of $i$-th table \\ 
% $\columnVar$ & column $k$ of $i$-th table \\ 
% $\question$ & the question \\
% $\qCons$ & $\ell$-th lexical constituent of $Q$ \\
% %$\qLen$ & Length of a question \\ 
% $\option$ & $m$-th answer option \\ 
% \cline{1-2}
% \bigstrut[t] $\xOne{.} $ & a unary variable \\
% $\xTwo{.}{.} $ & a pairwise variable \\
% %$ \interTableVar $ \\ 
% \hline
% \end{tabular}
% \caption{Notation for the ILP formulation. }
% \label{table:variables-and-description}
% \end{table}

\begin{table}[tb]
\centering 
\small
\begin{tabular}{r | L{30ex}} 
\multicolumn{2}{c}{\textsc{Basic Pairwise Activity Variables \bigstrut}} \\
\hline \hline 
$\xTwo{\tableCell}{\tableCellPrime} $ & \text{cell to cell} \\
$\xTwo{\tableCell}{\qCons} $ & cell to question constituent \\
$\xTwo{\header}{\qCons} $ & header to question constituent \\
$\xTwo{\tableCell}{\option}$ & cell to answer option \\ 
$\xTwo{\header}{\option}$ & header to answer option \\
$\xTwo{\columnVar}{\option}$ & column to answer option \\
$\xTwo{\tableVar}{\option}$ & table to answer option \\
$\xTwo{\columnVar}{\columnVarPrime}$ & column to column relation \\
\hline
\multicolumn{2}{c}{\textsc{High-level Unary Variables \bigstrut}}   \\
\hline \hline 
$\xOne{\tableVar}$ & active table \\ 
$\xOne{\rowVar}$  & active row \\
$\xOne{\columnVar} $ &  active column \\  
$\xOne{\header} $ & active column header  \\ 
$\xOne{\qCons}$ & active question constituent  \\ 
$ \xOne{\option} $ & active answer option \\ 
\hline
\end{tabular}
\caption{Variables used for defining the optimization problem for \tableilp\ solver. All variables have domain $\{0,1\}$.
%\tushar{Do we need column to answer option and table to answer option?} \daniel{Yea these last two are used in definition of constraints. }
}
\label{table:ilp-variables}
\end{table}

As previously mentioned, in practice we do not create all possible pairwise variables. Instead we choose the pairs alignment score $w(e)$ exceeds a pre-set threshold. For example, we create $\xTwo{\tableCell}{\tableCellPrimePrime}$ only if $w(\tableCell, \tableCellPrimePrime) \geq  \textsc{MinCellCellAlignment}$.\ifarxiv 
\footnote{An exhaustive list of the minimum alignment thresholds for creating pairwise variables is in Table~\ref{table:pairwise-thresholds} in the appendix. }   
\else
\footnote{An exhaustive list of the minimum alignment thresholds for creating pairwise variables is in Table~10
%\ref{table:pairwise-thresholds} 
of \cite{tableilp2016:arxiv}. }   
\fi 
% Table~\ref{table:ilp-variables} also includes some high level unary variables, which help conveniently impose structural constraints on the support graph $G$ we seek. An example is an \emph{active row} variable $\xOne{\tableVar}$ which takes value 1 if and only if at least a cell in row $j$ of table $i$ is in the support graph.
%\footnote{We say an element is ``active" if and only if the value of binary variable corresponding to this element is one.} 

%\textbf{Objective function: }
The objective function is a weighted linear sum over all variables instantiated for a given question answering problem.\ifarxiv 
\footnote{The complete list of weights for unary and pairwise variables is included in Table~\ref{table:objective-details} in the appendix.}
\else
\footnote{The complete list of weights for the pairwise and unary variables are included in Table~9
%\ref{table:objective-details} 
of \cite{tableilp2016:arxiv}.}
\fi 
%\footnote{There is a small set of auxiliary variables which are used for modeling complicated constraints, which will later introduce among constraints.}
%The weights for each The only detail about remaining to explain is how the weights for variables are set in the objective function (i.e. the vector $\textbf{w}$ in Eq. \ref{eq:ilp:obj}). 
A small set of auxiliary variables is defined for linearizing complicated constraints. %, which will later introduce among constraints.
% In addition to the current set of variables, there are  some auxiliary variables which we define when we talk about constraints. Defining auxiliary variables is a common trick for linearizing more intricate constraints at the cost of having more variables. 

Constraints are a significant part of our model, used for imposing the desired behavior on the support graph. 
%There is a long set of constraints defined in our model, 
Due to lack of space, we discuss only a representative subset here.\ifarxiv 
\footnote{The complete list of the constraints is explained in Table~\ref{table:constraints} in the appendix.}
\else
\footnote{The complete list of the constraints is explained in Table~13
%\ref{table:constraints}
of \cite{tableilp2016:arxiv}.}
\fi 
%Different groups of constraints are defined for different purposes, however it is hard to classify them into disjoint groups of constraints. We briefly mention a couple of important classes of constraints.

Some constraints relate variables to each other. For example, unary variables are defined through constraints that relate them to the corresponding pairwise variables. For instance, for active row variable $\xOne{\rowVar}$, we ensure that it is 1 if and only if at least one cell in row $j$ is active:
$$
\xOne{\rowVar}  \geq \xTwo{\tableCell}{*},\ \ \ \forall (\tableCell, *) \in \mathcal{R}_{ij}, \forall i, j, k, 
$$
where $\mathcal{R}_{ij}$ is collection of pairwise variables with one end in row  $j$ of table $i$. 
%(constraint~\ref{cons:rowIsActiveIfAnyCellInRowIsActive}, Table~\ref{table:constraints}). 

% Some constraints force the basic correctness principles on the final answer. For example $G$ should contain exactly one answer option which is expressed by the following: 
% $$
% \sum_{m}\xOne{\option} \leq 1, \quad  \sum_{m}\xOne{\option} \geq 1 
% $$
%constraint~\ref{cons:ifEdgeConnectedToTableIsActiveTableIsActive}, Table~\ref{table:constraints}. 
% Similarly $G$ should contain at least a certain number of elements in the question: 
% $$
%   \sum_{l} \xOne{\qCons} \geq \textsc{MinActiveQCons}  
% $$
%, which is modeled by constraint~\ref{cons:atLeastOneActiveEdgeConnectedToOption}, Table~\ref{table:constraints}.

% Another group of constraint induce simplicity (sparsity) in the output. For example $G$ should use at most a certain number of knowledge base table, since letting the inference use any table could lead to unreasonably long, and likely error-prone, answer chains. Also $G$ should use at most a certain number of cells in each table, since freedom to align with any number of cells in the table could lead to noisy and unreasonable alignments: 

In the remainder of this section, we outline some of the important characteristics we expect in our model, and provide details of a few illustrative constraints. 
}

\subsubsection{Basic Lookup}
Consider the following question:
\begin{myquote}
Which characteristic helps a fox find food?
(A) sense of smell (B) thick fur (C) long tail (D) pointed teeth
\end{myquote}
In order to answer such lookup-style questions, we generally seek a row with the highest aggregate alignment to question constituents. We achieve this by 
%boosting the objective function based on \peter{"based on" - vague! What does this mean?} the question-table alignment scores and the number of active question constituents. 
incorporating the question-table alignment variables with the alignment scores, $w(e)$, as coefficients and the active question constituents variable with a constant coefficient in the objective function. Since any additional question-table edge with a positive entailment score (even to irrelevant tables) in the support graph would result in an increase in the score, we disallow tables with alignments only to the question (or only to a choice) and add a small penalty for every table used in order to reduce noise in the support graph. We also limit the maximum number of alignments of a question constituent and \addedD{table cells, in order to prevent one constituent or cell from having a large influence on the objective function and thereby the solution: 
$$
\sum_{(*, \qCons) \in \mathcal{Q}_l } \xTwo{*}{\qCons} \leq \textsc{MaxAlignmentsPerQCons}, \forall l
$$
where $\mathcal{Q}_l$ is the set of all pairwise variables with one end in the question constituent $\ell$. 
}

\subsubsection{Parallel Evidence}
For certain questions, evidence needs to be combined from multiple rows of a table. For example,
\begin{myquote}
Sleet, rain, snow, and hail are forms of (A) erosion (B) evaporation (C) groundwater (D) precipitation
\end{myquote}
To answer this question, we need to combine evidence from multiple table entries from the weather terms table, \textit{(term, type)}, namely (sleet, precipitation), (rain, precipitation), (snow, precipitation), and (hail, precipitation). To achieve this, we allow multiple active rows in the support graph. Similar to the basic constraints, we limit the maximum number of active rows per table and add a penalty for every active row to ensure only relevant rows are considered for reasoning\addedD{:
$$
\sum_{j} \xOne{\rowVar} \leq \textsc{MaxRowsPerTable}, \forall i
$$
}
%Since our tables have well-defined schemas and few columns, 
To encourage only coherent parallel evidence within a single table,
we limit our support graph to always use the same columns across multiple rows within a table, i.e., every active row has \addedT{the active cells corresponding to the same  set of columns}.

\subsubsection{Evidence Chaining}
Questions requiring chaining of evidence from multiple tables, such as the example in Figure~\ref{fig:ny-example}, are typically the most challenging in this domain.
%especially for simple IR-based approaches. One such question was discussed in Section~\ref{sec:intro} which required chaining of tables about geographical locations, orbital events and their properties.
Chaining can be viewed as performing a \emph{join} between two tables. We introduce alignments between cells across columns %\tushar{I think this is resolved now, right ?} \daniel{Add ``conditioned on alignment of their corresponding header cell, "?}\tushar{It's now completely controlled by the allowed joins, right?}\ashish{It is; also Def.~\ref{def:support-graph} now explicitly already requires this}  
 in pairs of tables to allow for chaining of evidence. %\daniel{Now here we can again talk about red cells/edges in Fig 1, which are the active variables; also mention the green cells/edges which are used for alignment of the red stuff. How is that? }\tushar{That makes perfect sense. Also the same example question.} 
 To help minimize potential noise introduced by chaining irrelevant facts, we add a penalty for every inter-table alignment and also rely on the 0/1 weights of header-to-header edges
 %(shown in green in Figure \ref{fig:ny-example})
 to ensure only semantically meaningful table joins are considered.
%constrain the inter-table alignments to manually specified column pairs of similar type where the tables are from similar domains.  

\subsubsection{Semantic Relation Matching}
Our constraints so far have only looked at the content of the table cells, or the structure of the support graph, without explicitly considering the \emph{semantics} of the table schema. By using alignments between the question and column headers (i.e., type information), we exploit the table schema to prefer alignments to columns relevant to the ``topic'' of the question. In particular, for questions of the form ``which X $\ldots$'', we prefer answers that directly entail X or are connected to cells that entail X. However, this is not sufficient for questions such as:
\begin{myquote}
What is one way to change water \underline{from} a liquid \underline{to} a solid? (A) decrease the temperature (B) increase the temperature (C) decrease the mass (D) increase the mass
\end{myquote}
Even if we select the correct table, say $r_\text{change-init-fin}(c, i, f)$ that describes the initial and final states for a phase change event, both choice (A) and choice (B) would have the exact same score in the presence of table rows (increase temperature, solid, liquid) and (decrease temperature, liquid, solid). The table, however, does have the initial vs.\ final state structure. To capture this semantic structure, we annotate pairs of columns within certain tables with the semantic relationship present between them. In this example, we would annotate the phase change table with the relations: changeFrom$(c, i)$, changeTo$(c, f)$, and fromTo$(i, f)$. 

Given such semantic relations for table schemas, we can now impose a preference towards question-table alignments that respect these relations. We associate each semantic relation with a set of linguistic patterns describing how it might be expressed in natural language. \tableilp\ then uses these patterns to spot possible mentions of the relations in the question $Q$. We then add the soft constraint that for every pair of active columns in a table (with an annotated semantic relation) aligned to a pair of question constituents, there should be a valid expression of that relation in $Q$ between those constituents. In our example, we would match the relation fromTo(liquid, solid) in the table to ``liquid \underline{to a} solid'' in the question via the pattern ``X to a Y" associated with fromTo(X,Y), and thereby prefer aligning with the correct row (decrease temperature, liquid, solid).

\section{Evaluation}
\label{sec:experiments}

We compare our approach to three existing methods, demonstrating that it outperforms \addedA{the best previous} structured approach~\cite{aristo2015:mln} and produces a statistically significant improvement when used in combination with IR-based methods~\cite{aristo2016:combining}. For evaluations, we use a 2-core 2.5 GHz Amazon EC2 linux machine with 16 GB RAM.

\vspace{1ex}
\noindent \textbf{Question Set.}
\addedA{We use the same question set as \namecite{aristo2016:combining}, which consists of all non-diagram multiple-choice questions from 12 years of the NY Regents 4th Grade Science exams.\footnote{These are the only publicly available state-level science exams. http://www.nysedregents.org/Grade4/Science/home.html} The set is split into 108 development questions and 129 hidden test questions based on the year they appeared in (6 years each).} All numbers reported below are for the hidden test set, except for question perturbation experiments which relied on the 108 development questions.
%Although the low number of publicly released, real exam questions makes the dataset small, it provides sufficient signal for evaluation.

Test scores are reported as percentages. For each question, a solver gets a score of $1$ if it chooses the correct answer and $1/k$ if it reports a $k$-way tie that includes the correct answer. On the 129 test questions, a score difference of 9\% (or 7\%) is statistically significant at the 95\% (or 90\%, resp.) confidence interval based on the binomial exact test~\cite{howell2012statistical}.

%%%%%%%%%%%%%
%\subsection{Corpora, Rules, and Tables}
%
%\paragraph{Corpora}

\vspace{1ex}
\noindent \textbf{Corpora.}
We work with three knowledge corpora:
\begin{enumerate}
%\item Elementary Science Corpus: ~80k sentences about elementary science, consisting of a Regents study guide, CK12 textbooks,\footnote{www.ck12.org} and automatically collected Web sentences of similar style and content to that material.

\item Web Corpus: This corpus contains $5 \times 10^{10}$ tokens (280 GB of plain text) extracted from Web pages. \addedA{It was collected by Charles Clarke at the University of Waterloo, and has been \addedT{used previously by~\namecite{turney2013distributional}} and \namecite{aristo2016:combining}.} We use it here to compute statistical co-occurrence values for the \salience\ solver.

\item Sentence Corpus \addedA{\cite{aristo2016:combining}}: This includes sentences from the Web corpus above, as well as around 80,000 sentences from various domain-targeted sources for elementary science:
%to create a corpus of sentences for the IR-based solver. For the domain-targeted corpus, we collected 
a Regents study guide, CK12 textbooks (www.ck12.org), and web sentences with similar content as the course material.

\item Table Corpus (cf.~Section~\ref{subsec:tables}): This includes 65 tables totaling around 5,000 rows, designed based on the development set and study guides, as well as 4 Open IE-style~\cite{Banko2007OpenIE} automatically generated tables totaling around 2,600 rows.\footnote{\addedA{Table Corpus and the ILP model are available at allenai.org}.} 
%We also converted sentences from Regents study guide into OpenIE-style~\cite{etzioni2008open} tuples to create a 3-column table. 
%\tushar{Might be better to describe the tables (what extractor, etc) in more detail in the main section.}\tushar{Reduced the description here after I added more details in the main section.}

\end{enumerate}

%From the Elementary Science corpus, a rulebase of 45,000 rules was automatically extracted for \praline, and a datastore of 45 tables (total 10k rows, 40k cells) was built using an interactive table-building tool and manually, for \tableilp. The \salience\ solver uses the Web Corpus and a window of 10 words to compute PMI values.

%\daniel{Shall we add a little more explanation about the auto-generated tables? We need to refer to them in the ablation study. Here is what I wrote; see if are happy to fit it somewhere:  } \ashish{yes, should add a little about tables, hopefully without using the word `manually' much; ok to say manually defined schema.}\tushar{Maybe we don't need the Elementary science Corpus as we are referring to the OpenIE tuples here and we don't have RULE solver as a baseline.}
%\ashish{tushar: that makes sense. should we just have two corpora, the Web Corpus and the Table Corpus in the bullets above? Oh. do we need something for MLN or shd we just refer to the EMNLP implementation? probably the latter.}\tushar{I was thinking of just mentioning the MLN solver as the solver with its corpora. The tricky part is also the Lucene solver which has a mixture of both the corpora and more.}
%{\color{red}
%Some of the tables are automatically modeled as relations and OpenIE-style tuples~\cite{OPENIE} extracted from Barron's syllabus, using ``NameOfATool" (and some references here?). These tables help in bringing more coverage to the solver (compared to the manually generated tables, which were mostly based on the training data). }

%%%%%%%%%%%
\subsection{Solvers}
\label{section:baslines}

\vspace{1ex}
\noindent \textbf{\tableilp} (our approach). 
Given a question $Q$, we select the top 7 tables from the Table Corpus using the the standard TF-IDF score of $Q$ with tables treated as bag-of-words documents. For each selected table, we choose the 20 rows that overlap with $Q$ the most. This filtering improves efficiency and reduces noise. We then generate an ILP and solve it using the open source SCIP engine~\cite{solver:scip}, returning the active answer option $\option$ from the optimal solution. To check for ties, we disable $\option$, re-solve the ILP, \addedA{and compare the score of the second-best answer, if any, with that of $a_m$}.

%We compare our approach to three baseline solvers:
%Apart from three statistical baseline methods, we use the state-of-the-art structured method from prior work~\cite{aristo2015:mln} to compare against our approach

\vspace{1ex}
\noindent \textbf{MLN Solver} (structured inference baseline). 
We consider the current state-of-the-art structured reasoning method developed for this specific task by~\namecite{aristo2015:mln}. We compare against their best performing system, namely Praline, which uses Markov Logic Networks~\cite{richardson2006markov} to (a) align lexical elements of the question with probabilistic first-order science rules and (b) to control inference. We use \addedT{the entire set of} 47,000 science rules from their original work, which were also derived from same domain-targeted sources as the ones used in our Sentence Corpus.
%\daniel{This last phrase sounds odd to me: ``rule-based science knowledge, and to control inference.'' Anything missing from here?}

\vspace{1ex}
\noindent \textbf{\lucene\ Solver} (information retrieval baseline).
\addedT{We use the IR baseline by \namecite{aristo2016:combining},} \addedA{which selects the answer option that has} the best matching sentence in a corpus. \addedA{Specifically,} for each answer option $a_i$, the \lucene\ solver sends $q + a_i$ as a query to a search engine (we use Lucene) on the Sentence Corpus, and returns the search engine's score for the top retrieved sentence $s$, where $s$ must have at least one non-stopword overlap with $q$, and at least one with $a_i$. The option with the highest Lucene score is returned as the answer.

\vspace{1ex}
\noindent \textbf{\salience\ Solver} (statistical co-occurrence baseline).
\addedT{We use the PMI-based approach by \namecite{aristo2016:combining}}, \addedA{which selects} the answer option that most frequently co-occurs with the question words in a corpus. Specifically, it extracts unigrams, bigrams, trigrams, and skip-bigrams from the question and each answer option.
%We use the SMART stop word list \cite{salton1971smart} to filter the extracted n-grams, but allow trigrams to have a stop word in the middle.
For a pair $(x,y)$ of $n$-grams, their pointwise mutual information (PMI)~\cite{church1989} in the corpus is defined as $\log \frac{p(x,y)}{p(x)p(y)}$ where $p(x,y)$ is the co-occurrence frequency of $x$ and $y$ (within some window) in the corpus.
The solver returns the answer option that has the largest average PMI in the Web Corpus, calculated over all pairs of question $n$-grams and answer option $n$-grams.

%%%%%%%%%%%%%%%%%%%%%%%%
\subsection{Results}

We first compare the accuracy of our approach against the previous structured (MLN-based) reasoning solver. We also compare against \lucene(tables), an IR solver using table rows expressed as sentences, thus embodying an unstructured approach operating on the same knowledge as \tableilp.
%\daniel{cite \cite{howell2012statistical}?} \ashish{hmm. psychology :-) Is there a pure stats or CS reference?} \daniel{I tried to find another CS-ish reference, but couldn't find a significant highly cited reference. (This work is highly cited; beside hypothesis testing is the cornerstone of the psycology! :) )}
%\ashish{Howell it is, then!}
%\ashish{adding 90\% confidence interval too, to avoid people thinking nothing else besides one result is statistically significant} 

% \begin{table}[htb]
% \centering
% \setlength\tabcolsep{10pt}
% \setlength\doublerulesep{\arrayrulewidth}
% \begin{tabular}{c|l|c}
% %\textbf{Solver} & \textbf{Test score} \\
% Solver type & Solver & Test Score \\
% \hline\hline \bigstrut[t]
% \multirow{3}{*}{Unstructured} & \lucene (tables) & 51.2 \\
% & \lucene & 58.5 \\
% & \salience & 60.7 \\
% \hline \bigstrut[t]
% \multirow{2}{*}{Structured} \bigstrut[t]  % leave some gap (?)
% & \praline & 47.5 \\
% & \tableilp & {\bf 61.5} \\
% \hline
% \end{tabular}
% \caption{Test score comparison of the solvers using structured and unstructured data. \lucene(tables) uses the table rows as input sentences for the \lucene\ approach.}
% \label{tab:eval}
% \end{table}

\begin{table}[htb]
\centering
\small
\setlength\tabcolsep{10pt}
\setlength\doublerulesep{\arrayrulewidth}
\begin{tabular}{l|c}
Solver & Test Score (\%) \\
\hline\hline \bigstrut[t]
\praline & 47.5 \\
\lucene (tables) & 51.2 \\
\tableilp & {\bf 61.5} \\
\hline
\end{tabular}
\caption{\tableilp~significantly outperforms both the prior MLN reasoner, and IR using identical knowledge as \tableilp}
% \caption{Test score comparison: structured reasoning methods and IR using identical knowledge as \tableilp}
\label{tab:eval}
\end{table}

As Table \ref{tab:eval} shows, among the two structured inference approaches, \tableilp\ outperforms the MLN baseline by 14\%. \addedA{The preliminary ILP system reported by \namecite{aristo2016:combining} achieves only a score of 43.8\% on this question set.} Further, given the same semi-structured knowledge (i.e., the Table Corpus), \tableilp\ is substantially (+10\%) better at exploiting the structure than the \lucene(tables) baseline, \addedA{which, as mentioned above,} uses the same data expressed as sentences.
%Finally, \tableilp's score is comparable to the unstructured baselines that utilized much larger corpora. This closes, for the first time, the rather large gap between unstructured and structured approaches for this QA domain.

%\tableilp has significantly better performance compared to the IR baseline limited to the same knowledge base, although IR baseline gets very close to \tableilp when it uses significant amount of free data crawled from the web, as well as PMI baseline. \tableilp  has significantly better performance compared to a rival structured model MLN. 

\subsubsection{Complementary Strengths}

%Not all the solvers have the same behavior on the same problems; a question might be easy for one, meanwhile hard for the other. 
While their overall score is similar, \tableilp\ and IR-based methods clearly approach QA very differently. To assess whether \tableilp\ \addedA{adds} any new capabilities, we considered the 50 (out of 129) questions incorrectly answered by \salience\ solver (ignoring tied scores). On these unseen but arguably more difficult questions, \tableilp\ answered 27 questions correctly, achieving a score of 54\% compared to the random chance of 25\% for 4-way multiple-choice questions. Results with \lucene\ solver were similar: \tableilp\ scored 24.75 on the 52 questions incorrectly answered by \lucene\ (i.e., 47.6\% accuracy).
%For example, out of the 74 questions correctly answered by the \tableilp \ solver, \lucene\ solver also answers 50 of these questions and \salience \ answers 48 questions in the hidden set. 

\begin{table}[htb]
\centering
\small
\setlength\tabcolsep{10pt}
\setlength\doublerulesep{\arrayrulewidth}
\begin{tabular}{l|c}
%\textbf{Solver} & \textbf{Test score} \\
Solver & Test Score (\%) \\
\hline\hline \bigstrut[t]
\lucene & 58.5 \\
\salience & 60.7 \\
\tableilp & 61.5  \\
\tableilp\ + \lucene & 66.1 \\
\tableilp\ + \salience & 67.6 \\
\tableilp\ + \lucene + \salience & {\bf 69.0} \\
\hline
\end{tabular}
\caption{Solver combination results}
\label{tab:combination}
\end{table}

\addedA{This analysis highlights the} complementary strengths of these solvers. \addedA{Following \namecite{aristo2016:combining}, we create an ensemble of \tableilp, \lucene, and \salience\ solvers, combining their answer predictions} using a simple Logistic Regression model trained on the development set. \addedA{This model uses} 4 features derived from each solver's score for each answer option, and 11 features derived from \tableilp's support graphs.
\ifarxiv 
\footnote{Details of the 11 features may be found in the Appendix B.}
\else
\footnote{Details of the 11 features may be found in the extended version~\cite{tableilp2016:arxiv}.}
\fi Table \ref{tab:combination} shows the results,
%two top performing solvers from Table \ref{tab:eval}. 
%The addition of each solver to \tableilp\ leads to an increase in the score, 
with the final combination at 69\% representing a significant improvement over individual solvers.
%being statistically significantly better than \lucene\ alone. 
%\tushar{TODO: Use the statistics from the solver overlap to make this point stronger.}\tushar{This is taken care of by the earlier para, right?}

%%%%%%%%%%%%%%
\subsubsection{ILP Solution Properties}

Table~\ref{tab:stats} summarizes various ILP and support graph statistics for \tableilp, averaged across all test questions.

The optimization model has around 50 high-level constraints, which result, on average, in around 4000 inequalities over 1000 variables. Model creation, which includes %a large number of 
computing pairwise entailment scores using WordNet, takes 1.9 seconds on average per question, and the resulting ILP is solved by the SCIP engine in 2.1 seconds (total for all four options), using around 1,300 LP iterations for each option.\footnote{Commercial ILP solvers (e.g., CPLEX, Gurobi) are much faster than the open-source SCIP solver we used for evaluations.}
%\dan{would it be useful to say that there are much better solvers, that were not used only to keep to publicly avalialbe tools?}\ashish{possibly; does help with the scalability point}
Thus, \tableilp\ takes only 4 seconds to answer a question using multiple rows across multiple tables (typically 140 rows in total), as compared to 17 seconds needed by the \praline\ solver for reasoning with four rules (one per answer option).  

\begin{table}[htb]
\centering
\small
\setlength\tabcolsep{10pt}
\setlength\doublerulesep{\arrayrulewidth}
\begin{tabular}{llc}
Category & Quantity & Average \\
\hline\hline \bigstrut[t]
\multirow{3}{*}{ILP complexity} & \#variables & 1043.8 \\
& \#constraints & 4417.8 \\
& \#LP iterations & 1348.9 \\
\hline \bigstrut[t]
%\multirow{3}{*}{Alignments} &\#question constituents & 3.34 \\
%& \#rows & 2.32 \\
\multirow{2}{*}{Knowledge use} & \#rows & 2.3 \\
& \#tables & 1.3 \\
\hline \bigstrut[t]
\multirow{2}{*}{Timing stats} & model creation & 1.9 sec \\
& solving the ILP & 2.1 sec \\
\hline
\end{tabular}
\caption{\tableilp\ statistics averaged across questions}
\label{tab:stats}
\end{table}

While the final support graph on this question set relies mostly on a single table to answer the question, it generally combines information from more than two rows (2.3 on average) for reasoning. This suggests parallel evidence is more frequently used on this dataset than evidence chaining.
%\tableilp\ relies on roughly three question constituents to answer the questions which have XXX constituents on an average. \ashish{hmm. this doesn't sound that great and is probably one of the biggest weakness of the solver --- that it ignores a large part of the question! commenting it out until we have a better solution :-)}

%%%%%%%%%%%%%%%%%%%%%%%%
\subsection{Ablation Study}

To quantify the importance of various components of our system, we performed several ablation experiments, summarized in Table~\ref{tab:ablation} and described next.
% We start with two components related to structured inference, and then discuss the impact of coverage of knowledge and lexical alignment scores.
%\ashish{it's better to order these in terms of what's most relevant. Putting Lexical Entailment upfront can be troublesome, as we don't say much about it. Reordering.} \daniel{ok; fine by me.}

\begin{table}[htb]
\centering
\small
\setlength\tabcolsep{8pt}
\setlength\doublerulesep{\arrayrulewidth}
\begin{tabular}{ll|cc}
%\multicolumn{1}{c|}{\textbf{Approach}} & \textbf{Test score} & \textbf{\% drop} \\
\multicolumn{2}{c|}{Solver} & Test Score (\%) \\
\hline\hline
\multicolumn{2}{l|}{\bigstrut[t] \tableilp} & 61.5  \\
  & No Multiple Row Inference & 51.0 \\
  & No Relation Matching & 55.6 \\
  \cline{2-3} \bigstrut[t] 
  & No Open IE Tables & 52.3  \\
  & No Lexical Entailment & 50.5  \\
\hline 
\end{tabular}
\caption{Ablation results for \tableilp}
\label{tab:ablation}
\end{table}

\vspace{1ex}
\noindent \textbf{No Multiple Row Inference}: We modify the ILP constraints to limit inference to a single row (and hence a single table), thereby disallowing parallel evidence and evidence chaining (Section~\ref{subsec:ilp}). This drops the performance by 10.5\%, highlighting the importance of being able to combine evidence from multiple rows (which would correspond to multiple sentences in a corpus) from one or more tables.

\vspace{1ex}
\noindent \textbf{No Relation matching}: To assess the importance of considering the semantics of the table, we remove the requirement of matching the semantic relation present between columns of a table with its lexicalization in the question (Section~\ref{subsec:ilp}). The 6\% drop indicates \tableilp\ relies strongly on the table semantics to ensure creating meaningful inferential chains.
%gets similar performance to the IR-based solver on the table corpus, since it also picks the best matching sentences (corresponding to a row in a table) for a question.

\vspace{1ex}
\noindent \textbf{No Open IE tables}: To evaluate the impact of relatively unstructured knowledge from a large corpus, we removed the tables containing Open IE extractions (Section~\ref{subsec:qa-as-subgraph-selection}). The 9\% drop in the score shows that this knowledge is important and \tableilp\ is able to exploit it even though it has a very simple triple structure. This opens up the possibility of extending our approach to triples extracted from larger knowledge bases.

\vspace{1ex}
\noindent \textbf{No Lexical Entailment}: Finally, we test the effect of changing the alignment metric $w$ (Section~\ref{subsec:qa-as-subgraph-selection}) from WordNet based scores to a simple asymmetric word-overlap measured as $\mathit{score}(T, H) = \frac{|T \cap H|}{|H|}$. Relying on just word-matching results in an 11\% drop, which is consistent with our knowledge often being defined in terms of generalities.% and not overfit to the questions.

%\tushar{Daniel: Does my change look okay ? I have used most of your text with some changes.}
%There are interesting points can be inferred from the comparison in Table~\ref{tab:ablation}: (1) We test effect of replacing the entailment-based alignment metric \daniel{Better name? I don't want to use similarity metric since it implies symmetricity} with simple word-matching. Using entailment for connecting facts/cells helps significantly (compared to a naive similarity metrics). (2) Chaining facts (rather than looking up a fact in a single row)  is the second helping important component. (3) The auto-generated tables help the solver to generalize over the unseen test data, and hence the final performance. (4) The relation alignment adds a non-trivial amount to the final combination. 

%%%%%%%%%%%%%%%%%%%%%%%%
\subsection{Question Perturbation}
\label{subsec:perturbation}

One desirable property of QA systems is robustness to simple variations of a question, \emph{especially when a variation would make the question arguably easier for humans}. 
%\tushar{I like the point being made in the following line, but it highlights the limited nature of our test set :). is there a different way of saying it? -- A solver might get very good results on a limited dataset, but it is not clear if it generalizes well beyond that dataset. } \ashish{probably not. it's a sticky point. commenting it out.}
%
%In this experiment we test generalizability against a 

To assess this, we \addedA{consider a simple, automated way to perturb} each 4-way multiple-choice question: (1) query Microsoft's Bing search engine (www.bing.com) with the question text and obtain the text snippet of the top 2,000 hits; (2) create a list of strings by chunking and tokenizing the results; (3) remove stop words and special characters, as well as any words (or their lemma) appearing in the question; (4) sort the remaining strings based on their frequency; and (5) replace the three incorrect answer options in the question with the most frequently occurring strings, thereby generating a new question. For instance:
\begin{quote}
{In New York State, the longest period of daylight occurs during which month? (A) \emph{eastern} (B) June (C) \emph{history} (D) \emph{years}}
\end{quote}
%
%As can be seen here, the options (except for the correct one) are sometimes not even of the correct type, making it often \emph{easier} for humans. Surprisingly, such simple perturbations can often throw computers off. 
As in this example, the perturbations (italicized) are often not even of the correct ``type'', \addedA{typically making them much easier for humans. They, however, still remain difficult for solvers.}

\begin{table}[htb]
\centering
\small
\setlength\tabcolsep{10pt}
\setlength\doublerulesep{\arrayrulewidth}
\begin{tabular}{l|cccc}
 & Original & \multicolumn{2}{c}{\% Drop with Perturbation} \\
 \cline{3-4}
Solver & Score (\%) & \bigstrut[t] absolute & relative \\
\hline\hline \bigstrut[t]
\lucene & 70.7 & 13.8 & 19.5 \\
\salience & 73.6 & 24.4 & 33.2 \\
\tableilp & 85.0 & {\bf 10.5} & {\bf 12.3} \\
\hline
\end{tabular}
\caption{Drop in solver scores (on the development set, rather than the hidden test set) when questions are perturbed}
\label{tab:perturbation}
\end{table}

For each of the 108 development questions, we generate 10 new perturbed questions, using the 30 most frequently occurring words in step (5) above. \addedT{While this approach can introduce new answer options that should be considered correct as well, only 3\% of the questions in a random sample exhibited this behavior.} Table~\ref{tab:perturbation} shows the performance of various solvers on the resulting 1,080 perturbed questions. As one might expect, the \salience\ approach suffers the most at a 33\% relative drop. \tableilp's score drops as well (since answer type matching isn't perfect), but only by 12\%, attesting to its higher resilience to simple question variation.
%\dan{could you add a sentence explaining the reason for the drop? One motivation for this was the hope that type mismatch, for example, will never be supported...}
%every solvers' accuracy drops on this dataset. We observe, however, that \tableilp\ has the smallest percentage drop among other solvers. 

%\daniel{Shall we mention that this dataset is based on Regents-Train and has size 108*10? } \ashish{yes! we could give a hint that we obviously couldn't use unseen questions for this. so we took 108 other questions.}
%\ashish{also occurred to me that these 1080 questions are not ``independent'', so statistical significance doesn't quite directly apply even though we have a large set.}

%%%%%%%%%%%%%%%%%%%%%%%%
% \subsection{Error Analysis}

%\input{6.conclusion}

%%%%%%%%%%%%%% CONCLUSION %%%%%%%%%%%%%%

\section{Conclusion}
\label{sec:conclusion}

Answering real science questions is a challenging task because they are posed in natural language, require extensive domain knowledge, and often require combining multiple facts together.
We presented \tableilp, a system that can answer such questions, using a semi-structured knowledge base. We treat QA as a subgraph selection problem and then
formulate this as an ILP optimization. 
Most importantly, this formulation allows multiple, semi-formally expressed facts to be combined to answer questions, a capability outside the scope of IR-based QA systems.
In our experiments, this approach significantly outperforms both the previous best attempt at structured reasoning for this task, 
and an IR engine provided with the same knowledge. It also significantly boosts performance \addedA{when combined with} unstructured methods (IR and PMI). 
These results suggest that the approach is both viable and promising for natural language question answering.
\\
%\peter{Or a better last sentence? Or maybe speculate on next steps/ future work?}

\subsection*{Acknowledgments}
\addedD{ 
D.K. is in part supported by AI2 and Google. 
%We are indebted to Paul Allen whose long-term vision inspired this project, and continues to inspire our scientific endeavors. 
The authors would like to thank Christos Christodoulopoulos, Sujay Jauhar, Sam Skjonsberg, and the Aristo Team at AI2 for invaluable discussions and insights.  
}

%\ashish{changed named.bst to keep only initials and shorten months}
\bibliographystyle{mynamed}
\begin{small}
\bibliography{aristobib,extrabib}
\end{small}

\clearpage
\ifarxiv 
%%%%%%%%%%%%%% APPENDIX %%%%%%%%%%%%%%
\newpage

\appendix

\maketitle

\section{Appendix: ILP Model for \tableilp}

As it is widely known an ILP can be written as the following: 
\begin{align}
& \text{maximize}   && \mathbf{w}^\mathrm{T} \mathbf{x} \label{eq:ilp:obj}\\
& \text{subject to} && A \mathbf{x} \le \mathbf{b}, \label{eq:ilp:cons} \\
& \text{and} && \mathbf{x} \in \mathbb{Z}^n \label{eq:ilp:ints},
\end{align}
We first introduce the basic variables, and define the full definition of the ILP program: define the weights in the objective function ($\mathbf{w}$ in Equation \ref{eq:ilp:obj}), and the constraints ($A$ and $\mathbf{b}$ in Equation \ref{eq:ilp:cons}).

\textbf{Variables: }
We start with a brief overview of the basic variables and how they are combined into high level variables.

Table~\ref{table:variables-and-description} summarizes our notation to refer to various elements of the problem, such as $\tableCell$ for cell $(j,k)$ of table $i$, as defined in Section 3. We define variables over each element by overloading $\xOne{.}$ or $\xTwo{.}{.}$ notation which refer to a binary variable on elements or their pair, respectively. Table~\ref{table:ilp-variables} contains the complete list of basic variables in the model, all of which are binary. The pairwise variables are defined between pairs of elements; e.g., $\xTwo{\tableCell}{\qCons}$ takes value 1 if and only if the corresponding edge is present in the support graph. Similarly, if a node corresponding to an element of the problem is present in the support graph, we will refer to that element as being \emph{active}. 

\begin{table}[h]
\centering
\small
\begin{tabular}{|c|L{40ex}|} 
\hline
 \bigstrut[t] \textsc{Reference} & \bigstrut[t] \textsc{Description} \\ 
\hline
\bigstrut[t] $i$ & index over tables \\ 
$j$ & index over table rows \\ 
$k$ & index over table columns \\ 
$l$ & index over lexical constituents of question \\
$m$ & index over answer options \\
 \hline
% \bigstrut[t] $\tableVar$ & table $i$ \\ 
%$\tableLen$ & Number of tables \\ 
% $\header $  & header of the $k$-th column of $i$-th table \\
% $\tableCell $ & cell in row $j$ and column $k$ of $i$-th table \\ 
% $\rowVar$ & row $j$ of $i$-th table \\ 
% $\columnVar$ & column $k$ of $i$-th table \\ 
% $\question$ & the question \\
% $\qCons$ & $\ell$-th lexical constituent of $Q$ \\
%$\qLen$ & Length of a question \\ 
% $\option$ & $m$-th answer option \\ 
% \cline{1-2}
\bigstrut[t] $\xOne{.} $ & a unary variable \\
$\xTwo{.}{.} $ & a pairwise variable \\
%$ \interTableVar $ \\ 
\hline
\end{tabular}
\caption{Notation for the ILP formulation. }
\label{table:variables-and-description}
\end{table}

% \begin{table}
% \centering 
% \small
% \begin{tabular}{|r L{30ex}|} 
% \hline 
% \multicolumn{2}{|c|}{\textsc{Basic Pairwise Activity Variables \bigstrut}} \\
% $\xTwo{\tableCell}{\tableCellPrime} $ & \text{cell to cell} \\
% $\xTwo{\tableCell}{\qCons} $ & cell to question constituent \\
% $\xTwo{\header}{\qCons} $ & header to question constituent \\
% $\xTwo{\tableCell}{\option}$ & cell to answer option \\ 
% $\xTwo{\header}{\option}$ & header to answer option \\
% $\xTwo{\columnVar}{\option}$ & column to answer option \\
% $\xTwo{\tableVar}{\option}$ & table to answer option \\
% $\xTwo{\columnVar}{\columnVarPrime}$ & column to column relation \\
% \hline
% \multicolumn{2}{|c|}{\textsc{High-level Unary Variables \bigstrut}}   \\
% $\xOne{\tableVar}$ & active table \\ 
% $\xOne{\rowVar}$  & active row \\
% $\xOne{\columnVar} $ &  active column \\  
% $\xOne{\header} $ & active column header  \\ 
% $\xOne{\qCons}$ & active question constituent  \\ 
% $ \xOne{\option} $ & active answer option \\ 
% \hline
% \end{tabular}
% \caption{Variables used for defining the optimization problem for \tableilp\ solver. All variables have domain $\{0,1\}$.
% }
% \label{table:ilp-variables}
% \end{table}

In practice we do not create pairwise variables for all possible pairs of elements; instead we create pairwise variables \addedT{for edges that have an entailment score exceeding a threshold}. For example we create the pairwise variables $\xTwo{\tableCell}{\tableCellPrimePrime}$ only if $w(\tableCell, \tableCellPrimePrime) \geq  \textsc{MinCellCellAlignment}$. An exhaustive list of the minimum alignment thresholds for creating pairwise variables is in Table~\ref{table:pairwise-thresholds}.   

Table~\ref{table:ilp-variables} also includes some high level unary variables, which help conveniently impose structural constraints on the support graph $G$ we seek. An example is the \emph{active row} variable $\xOne{\tableVar}$ which should take value 1 if and only if at least a cell in row $j$ of table $i$.
%\footnote{We say an element is ``active" if and only if the value of binary variable corresponding to this element is one.} 

\begin{table*} 
\centering
\small 
\renewcommand{\sfdefault}{phv}
\resizebox{\textwidth}{!}
{
\begin{tabular}{|l|lc|lc|lc|lc|}	
    \hline 
	\multirow{2}{*}{Pairwise Variables} &
	% \parbox[t]{2mm}{\multirow{2}{*}{\rotatebox[origin=c]{90}{}}} &
 	$\xTwo{\tableCell}{\tableCellPrimePrime} $ & 1 &
    $\xTwo{\tableCell}{\tableCellPrime} $ & $w(\tableCell, \tableCellPrime) - 0.1$ &
    $\xTwo{\tableCell}{\qCons} $ &  $w(\qCons, \tableCell)$ &  
    $\xTwo{\header}{\qCons} $ &  $w(\qCons, \header)$  \\   
    & $\xTwo{\tableCell}{\option}$ & $w(\tableCell, \option)$ & 
    $\xTwo{\header}{\option}$ & $w(\header, \option)$ & 
    %$\xTwo{\columnVar}{\columnVarPrime}$ & {\color{red} ?} 
    &
    & & \\   
    \hline 
 	\multirow{2}{*}{Unary Variables}
    & $\xOne{\tableVar}$ & 1.0  &  
      $\xOne{\rowVar}$   & -1.0 & 
      $\xOne{\columnVar} $  & 1.0 &   
      $\xOne{\header} $ & 0.3 \\ 
      & $\xOne{\tableCell} $  & 0.0 &    
      $\xOne{\qCons}$ & 0.3 &   
      & & & \\
	\hline 
%     \multirow{2}{*}{Other Auxilary Variables} & $\xOne{\text{whichTermIsActive}}$ & 1.5d  &   $\xOne{\text{whichTermIsAligned}}$ &1.5d & 
% $\xOne{\text{cellProximityBoost}, i_1, i_2}$ & $1 / (i_1 - i_2 + 1)$& & & \\
%     & & & & & & & & & \\
%     \hline 
\end{tabular}
}
\caption{The weights of the variables in our objective function. In each column, the weight of the variable is mentioned on its right side. The variables that are not mentioned here are set to have zero weight.}
\label{table:objective-details}
\end{table*}

\textbf{Objective function: }
Any of the binary variables defined in our problem are included in the final weighted linear objective function. The weights of the variables in the objective function (i.e. the vector $\textbf{w}$ in Equation \ref{eq:ilp:obj}) are \addedT{set} according to Table~\ref{table:objective-details}. In addition to the current set of variables, \addedT{we introduce auxiliary variables for certain constraints}. Defining auxiliary variables is a common trick for linearizing more intricate constraints at the cost of having more variables. 

\textbf{Constraints: }
Constraints are significant part of our model in imposing the desirable behaviors for the \textit{support graph} (cf. Section 3.1). 

The complete list of the constraints is explained in Table~\ref{table:constraints}. While groups of constraints are defined for different purposes, it is hard to partition them into disjoint sets of constraints. Here we give examples of some important constraint groups. 

\textbf{Active variable constraints:} An important group of constraints relate variables to each other. The unary variables are defined through constraints that relate them to the basic pairwise variables. For example, active row variable $\xOne{\tableVar}$ should be active if and only if any cell in row $j$ is active.  (constraint~\ref{cons:rowIsActiveIfAnyCellInRowIsActive}, Table~\ref{table:constraints}).

\textbf{Correctness Constraints:} A simple, but important set of constraints force the basic correctness principles on the final answer. For example $G$ should contain exactly one answer option which is expressed by constraint~\ref{cons:OnlyASingleOption}, Table~\ref{table:constraints}. Another example is that, $G$ should contain at least a certain number of constituents in the question, which is modeled by constraint~\ref{eq:MinActiveQCons}, Table~\ref{table:constraints}.
% \noindent \textbf{Correctness:}
% \begin{itemize}
%	\item The nodes of the proof graph $\mathcal{G}$ consist of table cells, question constituents or question options. 
%    \item The proof graph $\mathcal{G}$ needs to be a DAG with its ends in both the question constituents and its options. The combination of the constraints we introduce here satisfy a DAG. 
%     \item $G$ should contain exactly one answer option (constraint~\ref{cons:ifEdgeConnectedToTableIsActiveTableIsActive}, Table~\ref{table:constraints}). 
%     \item $G$ should contain at least a certain number of elements in the question (constraint~\ref{cons:atLeastOneActiveEdgeConnectedToOption}, Table~\ref{table:constraints}).
%     \item If a table is used, at least some of its columns should be active as well (constraint~\ref{cons:ifQConsIsActiveAtLeastOneEdgeConnectedToIt}, Table~\ref{table:constraints}). 
%     \item If a table is used, at least some of its columns should be active as well (constraint~\ref{cons:OnlyASingleOption}, Table~\ref{table:constraints}).
%     \item If a cell is used, the sum of weights of all alignments to it must be at least a certain value (constraint~\ref{cons:aCellIsActiveIffTheSumOfExtAlignIsAtLeastSth}, Table~\ref{table:constraints}).
% \end{itemize}

\textbf{Sparsity Constraints:} Another group of constraint induce simplicity (sparsity) in the output. For example $G$ should use at most a certain number of knowledge base tables (constraint~\ref{eq:MAXTABLESTOCHAIN}, Table~\ref{table:constraints}), since letting the inference use any table could lead to unreasonably long, and likely error-prone, answer chains.

%Table~\ref{table:constraints}. 
%Also $G$ should use at most a certain number of cells in each table (constraint~\ref{cons:ifEdgeConnectedToConsConsMustBeActive}, Table~\ref{table:constraints}); freedom to align with any number of cells in the table could lead to noisy and unreasonable alignments. 

% \noindent \textbf{Simplicity:}
% \begin{itemize}
%     \item $G$ should use at most a certain number of rows from each table (constraint~\ref{cons:optionShouldBeActiveIfAnyEdgeConnectedToItIsActive}, Table~\ref{table:constraints}). 
%     \item $G$ should use at most a certain number of knowledge base table (constraint~\ref{cons:ifTableIsActiveAtLeastAnEdgeIsConnctedToIt}, Table~\ref{table:constraints}). Letting the inference use any table could lead to unreasonably long, and likely error-prone, answer chains. 
%     \item $G$ should use at most a certain number of cells in each table (constraint~\ref{cons:ifEdgeConnectedToConsConsMustBeActive}, Table~\ref{table:constraints}). Freedom to align with any number of cells in the table could lead to noisy and unreasonable alignments.
% \end{itemize}

\section {Appendix: Features in Solver Combination}

\addedT{To combine the predictions from all the solvers, we learn a Logistic Regression model~\cite{aristo2016:combining} that returns a probability for an answer option, $a_i$, being correct based on the following features.}

\textbf{Solver-independent features:}
Given the solver scores $s_j$ for all the answer options $j$, we generate the following set of features for the answer option $a_i$, for each of the solvers:
\begin{enumerate}
\item Score = $s_i$
\item Normalized score = $\frac{s_i}{\sum_j s_j}$
\item Softmax score = $\frac{\exp(s_i)}{\sum_j \exp(s_j)}$
\item Best Option, set to $1$ if this is the top-scoring option = $\mathbb{I}(s_i = \max s_j)$ 
\end{enumerate}

\textbf{TableILP-specific features:}
Given the proof graph returned for an option, we generate the following 11 features apart from the solver-independent features:
\begin{enumerate}
\item Average alignment score for question constituents
\item Minimum alignment score for question constituents
\item Number of active question constituents
\item Fraction of active question constituents
\item Average alignment scores for question choice
\item Sum of alignment scores for question choice
\item Number of active table cells
\item Average alignment scores across all the edges
\item Minimum alignment scores across all the edges
\item Log of number of variables in the ILP
\item Log of number of constraints in the ILP
\end{enumerate}

\begin{table*}
\centering
\small 
\renewcommand{\sfdefault}{phv}
\resizebox{\textwidth}{!}{
\begin{tabular}{|lc|lc|lc|}	
    \hline 
    \textsc{MinCellCellAlignment} &  0.6 & \textsc{MinCellQConsAlignment} & 0.1  & \textsc{MinTitleQConsAlignment} & 0.1 \\ 
    \textsc{MinTitleTitleAlignment} &  0.0 & \textsc{MinCellQChoiceAlignment} & 0.2 & \textsc{MinTitleQChoiceAlignment} & 0.2\\ 
    \textsc{MinCellQChoiceConsAlignment} &  0.4 & \textsc{MinCellQChoiceConsAlignment} & 0.4 & \textsc{MinTitleQChoiceConsAlignment} & 0.4 \\ 
    \textsc{MinActiveCellAggrAlignment} & 0.1 & \textsc{MinActiveTitleAggrAlignment} & 0.1&
 &   \\ 
\hline 
\end{tabular}
}
\caption{Minimum thresholds used in creating pairwise variables.  }
\label{table:pairwise-thresholds}
\end{table*}

\begin{table*}
\centering
\small 
\renewcommand{\sfdefault}{phv}
\resizebox{\textwidth}{!}{
\begin{tabular}{|lc|lc|lc|}	
    \hline 
    \textsc{MaxTablesToChain} &  4 & \textsc{qConsCoalignMaxDist} & 4 & \textsc{WhichTermSpan} &  2 \\ 
%     \textsc{MinCellCellAlignment} &  0.6 & \textsc{MinCellQConsAlignment} & 0.1  & \textsc{MinTitleQConsAlignment} & 0.1 \\ 
%     \textsc{MinTitleTitleAlignment} &  0.0 & \textsc{MinCellQChoiceAlignment} & 0.2 & \textsc{MinTitleQChoiceAlignment} & 0.2\\ 
%     \textsc{MinCellQChoiceConsAlignment} &  0.4 & \textsc{MinCellQChoiceConsAlignment} & 0.4 & \textsc{MinTitleQChoiceConsAlignment} & 0.4 \\ 
%     \textsc{MinActiveCellAggrAlignment} & 0.1 & \textsc{MinActiveTitleAggrAlignment} & 0.1& \textsc{MinActiveTitleAggrAlignment} & 0.1 \\ 
%     \textsc{ActiveCellObjCoeff} &  0 & \textsc{ActiveCellObjCoeff} &  0 & \textsc{ActiveColObjCoeff} & 1 \\ 
%     \textsc{ActiveTitleObjCoeff} &  0.3 & \textsc{ActiveTitleObjCoeff} & 1 & \textsc{TableScoreObjCoeff} & 1 \\ 
%     \textsc{ActiveQConsObjCoeff} &  0.3 & \textsc{ActiveScienceTermBoost} & 2 & \textsc{WhichTermAddBoost} & 1.5 \\ 
    \textsc{WhichTermMulBoost} &  1 & \textsc{MinAlignmentWhichTerm} & 0.6 & \textsc{TableUsagePenalty} & 3 \\ 
    \textsc{RowUsagePenalty} &  1 & \textsc{InterTableAlignmentPenalty} & 0.1 & \textsc{MaxAlignmentsPerQCons} & 2 \\ 
    \textsc{MaxAlignmentsPerCell} &  2 & \textsc{RelationMatchCoeff} & 0.2 & \textsc{RelationMatchCoeff} & 0.2 \\ 
     \textsc{EmptyRelationMatchCoeff} & 0.0 & \textsc{NoRelationMatchCoeff} & -5 & \textsc{MaxRowsPerTable}  & 4 \\ 
     \textsc{MinActiveQCons} & 1 & \textsc{MaxActiveColumnChoiceAlignments} & 1 & \textsc{MaxActiveChoiceColumnVars} & 2 \\
\textsc{MinActiveCellsPerRow} & 2 & & & & \\ 
\hline 
\end{tabular}
}
\caption{Some of the important constants and their values in our model.  }
\label{table:coefficients}
\end{table*}

\newpage 
\newcommand{\linespacingInTable}{0.6}

\begin{table*}
	\small 
	\renewcommand{\arraystretch}{0.0}% Tighter
    \linespread{\linespacingInTable}\selectfont\centering
    %% fit the page 
    \resizebox{\textwidth}{!}{
	\begin{tabular}{|L{7.5cm}L{9.5cm}|}
      \hline 
      Collection of basic variables connected to header column  $k$ of table  $i$: & 
      \begin{equation}
          \mathcal{H}_{ik} = \setOf{(\header, \qCons); \forall l } \cup \setOf{(\header, \option); \forall m} 
      \end{equation} \\
            Collection of basic variables connected to cell $j, k$ of table $i$: & 
       \begin{equation}
              \mathcal{E}_{ijk} = \setOf{(\tableCell, \tableCellPrime); \forall i', j', k' } \cup \setOf{(\tableCell, \option); \forall m} \cup \setOf{(\tableCell, \qCons); \forall l}  
      \end{equation} \\
            Collection of basic variables connected to column  $k$  of table  $i$ & 
       \begin{equation}
               \mathcal{C}_{ik} = \mathcal{H}_{ik} \cup \left(  \bigcup_{j} \mathcal{E}_{ijk} \right)
      \end{equation} \\
            Collection of basic variables connected to row  $j$  of table  $i$: & 
       \begin{equation}
          \mathcal{R}_{ij} =  \bigcup_{k} \mathcal{E}_{ijk} 
      \end{equation} \\
            Collection of non-choice basic variables connected to row  $j$  of table  $i$:  & 
       \begin{equation}
          \mathcal{L}_{ij} = \setOf{(\tableCell,\tableCellPrime); \forall k, i', j', k'} \cup \setOf{ (\tableCell, \qCons); \forall k, l} 
      \end{equation} \\
            Collection of non-question basic variables connected to row  $j$  of table  $i$:   & 
       \begin{equation}
          \mathcal{K}_{ij} = \setOf{(\tableCell,\tableCellPrime); \forall k, i', j', k'} \cup \setOf{ (\tableCell, \option); \forall k, m}  
      \end{equation} \\
            Collection of basic variables connected to table  $i$:  & 
       \begin{equation}
              \mathcal{T}_i = \bigcup_{k} \mathcal{C}_{ik}
      \end{equation} \\
            Collection of non-choice basic variables connected to table  $i$:  & 
       \begin{equation}
              \mathcal{N}_{i} = \setOf{(\header, \qCons); \forall l } \cup \setOf{(\tableCell,\tableCellPrime); \forall j, k, i', j', k'}  \cup \setOf{ (\tableCell, \qCons); \forall j, k, l}  
      \end{equation} \\
            Collection of basic variables connected to question constituent $\qCons$:    & 
       \begin{equation}
              \mathcal{Q}_l = \setOf{ (\tableCell, \qCons); \forall i, j, k}  \cup \setOf{ (\header, \qCons); \forall i, k}  
      \end{equation} \\
            Collection of basic variables connected to option  $m$ & 
       \begin{equation}
              \mathcal{O}_m = \setOf{ (\tableCell, \option); \forall i, j, k}  \cup \setOf{ (\header, \option); \forall i, k}  
      \end{equation} \\
            Collection of basic variables in column $k$ of table $i$ connected to option  $m$: & 
       \begin{equation}
              \mathcal{M}_{i, k, m} = \setOf{ (\tableCell, \option); \forall j}  \cup \setOf{ (\header, \option)}  
      \end{equation} \\
      \hline 
    \end{tabular}
    }
  \caption{All the sets useful in definitions of the constraints in Table~\ref{table:constraints}. }
  \label{table:ilp-sets}
\end{table*}

\newpage

\begin{table*}
	\small 
	\renewcommand{\arraystretch}{0.0}% Tighter
    \linespread{\linespacingInTable}\selectfont\centering
    \resizebox{\textwidth}{!}{
	\begin{tabular}{|L{9.5cm}L{7.5cm}|}
      \hline 
		If any cell in row $j$ of table $i$ is active, the row should be active.
        & 
      \begin{equation}
      	  \label{cons:rowIsActiveIfAnyCellInRowIsActive}
          \xOne{\rowVar}  \geq \xTwo{\tableCell}{e}, \forall (\tableCell, e) \in \mathcal{R}_{ij}, \forall i, j, k 
      \end{equation} \\
      	If the row $j$ of table $i$ is active, at least one cell in that row must be active as well. 
        & 
      \begin{equation}
          \sum_{(\tableCell, e) \in \mathcal{R}_{ij}} \xTwo{\tableCell}{e}  \geq  \xOne{\rowVar}, \forall i, j
      \end{equation} \\
      	 Column $j$ header should be active if any of the basic variables with one end in this column header are active. 
        & 
      \begin{equation}
          \xOne{\header}  \geq \xTwo{\header}{e}, \forall (\header, e) \in \mathcal{H}_{ik}, \forall i, k
      \end{equation} 
      \\
            	If the header of column $j$ variable is active, at least one basic variable with one end in the end in the header
        & 
      \begin{equation}
          \sum_{(\header, e) \in \mathcal{H}_{ik}} \xTwo{\header}{e}  \geq  \xOne{\header}, \forall i 
      \end{equation} \\
		Column $k$ is active if at least one of the basic variables with one end in this column are active. 
        & 
      \begin{equation}
          \xOne{\columnVar} \geq \xTwo{\tableCell}{e}, \forall (\tableCell, e) \in \mathcal{C}_{ik}, \forall i, k
      \end{equation} \\
            \hline
    \end{tabular}
    }
\end{table*}

\begin{table*}
	\small 
	\renewcommand{\arraystretch}{0.0}% Tighter
 \linespread{\linespacingInTable}\selectfont\centering
    \resizebox{\textwidth}{!}{
	\begin{tabular}{|L{9.5cm}L{7.5cm}|}
          \hline 
                	If the column $k$ is active, at least one of the basic variables with one end in this column should be active. 
        & 
      \begin{equation}
          \sum_{(\tableCell, e) \in \mathcal{C}_{ik}} \xTwo{\tableCell}{e} \geq  \xOne{\header}, \forall i, k   
      \end{equation} \\
      	If a basic variable with one end in table $i$ is active, the table variable is active. 
        & 
      \begin{equation}
      		\label{cons:ifEdgeConnectedToTableIsActiveTableIsActive}
          \xTwo{\tableCell}{e} \geq \xOne{\tableVar}, \forall (\tableCell, e) \in \mathcal{T}_i, \forall i 
      \end{equation} \\
         If the table $i$ is active, at least one of the basic variables with one end in the table should be active.  
        & 
      \begin{equation}
      		\label{cons:ifTableIsActiveAtLeastAnEdgeIsConnctedToIt}
          \sum_{(t, e) \in \mathcal{T}_i} \xTwo{t}{e} \geq \xOne{\tableVar}, \forall i 
      \end{equation} \\
	  	If any of the basic variables with one end in option $\option$ are on, the option should be active as well.  
        & 
      \begin{equation}
      		\label{cons:optionShouldBeActiveIfAnyEdgeConnectedToItIsActive}
          \xOne{\option} \geq \xTwo{x}{\option}, \forall (e, \option) \in \mathcal{O}_m  
      \end{equation} \\
          If the question option $\option$ is active, there is at least one active basic element connected to it 
        & 
      \begin{equation}
      		\label{cons:atLeastOneActiveEdgeConnectedToOption}
          \sum_{(e, a) \in \mathcal{O}_m} \xTwo{x}{a} \geq  \xOne{\option}   
      \end{equation} \\
          If any of the basic variables with one end in the constituent $\qCons$, the constituent must be active.  
        & 
      \begin{equation}
      		\label{cons:ifEdgeConnectedToConsConsMustBeActive}
          \xOne{\qCons} \geq \xTwo{e}{\qCons}, \forall (e, \qCons) \in \mathcal{Q}_l  
      \end{equation} \\
          If the constituent $\qCons$ is active, at least one basic variable connected to it must be active. 
        & 
      \begin{equation}
      		\label{cons:ifQConsIsActiveAtLeastOneEdgeConnectedToIt}
          \sum_{(e, \qCons) \in \mathcal{Q}_l} \xTwo{e}{\qCons}  \geq  \xOne{\qCons}   
      \end{equation} \\
          Choose only a single option. 
        & 
      \begin{equation}
      	  \label{cons:OnlyASingleOption}
          \sum_{m}\xOne{\option} \leq 1, \quad  \sum_{m}\xOne{\option} \geq 1 
      \end{equation} \\
           There is an upper-bound on the number of active tables; this is to limit the solver and reduce the chance of using spurious tables.  
        & 
      \begin{equation}
      		\label{eq:MAXTABLESTOCHAIN}
          \sum_{i} \xOne{\tableVar} \leq \textsc{MaxTablesToChain}  
      \end{equation} \\
          The number of active rows in each table is upper-bounded. 
        & 
      \begin{equation}
          \sum_{j} \xOne{\rowVar} \leq \textsc{MaxRowsPerTable}, \forall i
      \end{equation} \\
          The number of active constituents in each question is lower-bounded. Clearly We need to use the question definition in order to answer a question.  
        & 
      \begin{equation}
          \sum_{l} \xOne{\qCons} \geq \textsc{MinActiveQCons}  
          \label{eq:MinActiveQCons}
      \end{equation} \\
%           Most number of cells in each table 
%         & 
%       \begin{equation}
%           \forall m \sum_{j, k} \xTwo{\tableCell}{ \option } \geq \textsc{MinActiveCell}  
%       \end{equation} \\
%           If a column header is active, some of the cells in the same column should be active as well  
%         & 
%       \begin{equation}
%           \forall i, k, \sum_{j} \xOne{\tableCell} \geq \xOne{\header} 
%       \end{equation} \\
          A cell is active if and only if the sum of coefficients of all external alignment to it is at least a minimum specified value 
        & 
        \begin{multline}
           \label{cons:aCellIsActiveIffTheSumOfExtAlignIsAtLeastSth}
          \sum_{(\tableCell, e) \in \mathcal{E}_{i, j, k}}  \xTwo{\tableCell}{e}  \geq \xOne{\tableCell} \\ 
          \times \textsc{MinActiveCellAggrAlignment}, \forall i,j, k 
		\end{multline} \\
          A title is active if and only if the sum of coefficients of all external alignment to it is at least a minimum specified value
        & 
        \begin{multline}
          \sum_{(\title, e) \in \mathcal{H}_{i, k}} \xTwo{\tableCell}{e} \geq \xOne{\tableCell} \\  \times \textsc{MinActiveTitleAggrAlignment}, \forall i, k
      	\end{multline} \\
         If a column is active, at least one of its cells must be active as well. & 
      \begin{equation}
          \sum_{j} \xOne{\tableCell} \geq \xOne{\columnVar}, \forall i, k
      \end{equation} \\
          At most a certain number of columns can be active for a single option 
 & 
      \begin{multline}
          \sum_{k} \xTwo{\columnVar}{\option}  \leq \textsc{MaxActiveChoiceColumn}, \\ \forall i, m
      \end{multline} \\
          If a column is active for a choice, the table is active too.  & 
      \begin{equation}
          	\xOne{\columnVar} \leq \xOne{\tableVar}, \forall i,k
      \end{equation} \\
          If a table is active for a choice, there must exist an active column for choice. & 
      \begin{equation}
          \xOne{\tableVar} \leq \sum_{k}  \xOne{\columnVar}, \forall i
      \end{equation} \\
          If a table is active for a choice, there must be some non-choice alignment.  & 
      \begin{equation}
          \xTwo{\tableVar}{\option} \leq \sum_{(e, e') \in \mathcal{N}_{i}} \xTwo{e}{e'}, \forall i, m
      \end{equation} \\
                 Answer should be present in at most a certain number of tables & 
      \begin{multline}
          \xTwo{\tableVar}{\option}  \leq \textsc{MaxActiveTableChoiceAlignmets}, \\ \forall i, m
      \end{multline} \\
          If a cell in a column, or its header is aligned with a question option, the column is active for question option as well.  & 
      \begin{multline}
          \xTwo{\tableCell}{\option}  \leq \xTwo{\columnVar}{\option}, \\ \forall i, k, m, \forall (\tableCell, \option) \in \mathcal{M}_{i, k, m}
      \end{multline} \\
          If a column is active for an option, there must exist an alignment to header or cell in the column. 	 & 
      \begin{equation}
          \xTwo{\columnVar}{\option} \leq \sum_{(\tableCell, \option) \in \mathcal{O}_{i, k, m}} \xTwo{\tableCell}{\option}, \forall i, m
      \end{equation} \\
      \hline 
    \end{tabular}
    }
\end{table*}

\begin{table*}
	\small 
	\renewcommand{\arraystretch}{0.0}% Tighter
    \linespread{\linespacingInTable}\selectfont\centering
    %% fit the page 
    \resizebox{\textwidth}{!}{
	\begin{tabular}{|L{9.0cm}L{8cm}|}
      \hline 
           At most a certain number of columns may be active for question option in a table. & 
      \begin{multline}
          \sum_{k} \xTwo{\columnVar}{\option} \leq \\ \textsc{MaxActiveChoiceColumnVars}, \forall i, m
      \end{multline} \\
          If a column is active for a choice, the table is active for an option as well. & 
      \begin{equation}
          \xTwo{\columnVar}{\option} \leq \xTwo{\tableVar}{\option}, \forall i, k, m
      \end{equation} \\
          If the table is active for an option, at least one column is active for a choice & 
      \begin{equation}
      \xTwo{\tableVar}{\option} \leq \sum_{k} \xTwo{\columnVar}{\option}, \forall i, m
      \end{equation} \\
      	 Create an auxiliary variable $\xOne{\text{whichTermIsActive}}$ with objective weight 1.5 and activate it, if there a ``which" term in the question. 
         & 
      \begin{multline}
          \sum_{l} 1\setOf{\qCons = \text{``which''} } \leq \xOne{\text{whichTermIsActive}}
      \end{multline} \\      
      Create an auxiliary variable $\xOne{\text{whichTermIsAligned}}$ with objective weight 1.5. Add a boost if at least one of the table cells/title aligning to the choice happens to have a good alignment ($\setOf{w(., .) > \textsc{MinAlignmentWhichTerm}}$) with the ``which'' terms, i.e. \textsc{WhichTermSpan} constituents after ``which''. 
         & 
      \begin{multline}
          \sum_i \sum_{(e_1, e_2) \in \mathcal{T}_i} \xTwo{e_1}{e_2} %1 \setOf{w(\text{``which''}, e_1) > \textsc{MinAlignmentWhichTerm}} 
            \geq  \xOne{\text{whichTermIsAligned}}
      \end{multline} \\
      	 A question constituent may not align to more than a certain number of cells & 
      \begin{equation}
          \sum_{(e, \qCons) \in \mathcal{Q}_l } \xTwo{e}{\qCons} \leq \textsc{MaxAlignmentsPerQCons}
      \end{equation} \\
      	  Disallow aligning a cell to two question constituents if they are too far apart; in other words add the following constraint if the two constituents $\qCons$ and $\qConsPrime$ are more than \textsc{qConsCoalignMaxDist} apart from each other: & 
      \begin{equation}
          \xTwo{\tableCell}{\qCons} + \xTwo{\tableCell}{\qConsPrime} \leq 1,  \forall l, l', i, j, k
      \end{equation} \\
      	 For any two two question constraints that are not more than \textsc{qConsCoalignMaxDist} apart create an auxiliary binary variable $\xOne{\text{cellProximityBoost}}$ and set its weight in the objective function to be $1 / (l - l' + 1)$, where $l$ and $l'$ are the indices of the two question constituents. With this we boost objective score if a cell aligns to two question constituents that are within a few words of each other & 
      \begin{multline}
          \xOne{\text{cellProximityBoost}} \leq \xTwo{\tableCell}{\qCons}, \\ \xOne{\text{cellProximityBoost}} \leq  \xTwo{\tableCell}{\qConsPrime}, \forall i, j, k 
      \end{multline} \\
      If a relation match is active, both the columns for the relation must be active & 
      \begin{equation}
      	 \label{cons:IfRelationIsActiveBothColumnsMustBeActive}
         \xFour{\columnVar}{\columnVarPrime}{\qCons}{\qConsPrime} \leq \xOne{\columnVar}, \xFour{\columnVar}{\columnVarPrime}{\qCons}{\qConsPrime} \leq \xOne{\columnVarPrime}
      \end{equation} \\
       If a column is active, a relation match connecting to the column must be active & 
      \begin{equation}	\label{cons:ifColumnIsActiveARelationConnectingToItMustBeActive}
        \xOne{\columnVar} \leq \sum_{k'} (\xFour{\columnVar}{\columnVarPrime}{\qCons}{\qConsPrime} + \xFour{\columnVarPrime}{\columnVar}{\qCons}{\qConsPrime}), \forall k  
      \end{equation} \\
      	  If a relation match is active, the column cannot align to the question in an invalid position & 
      \begin{multline}
 \label{cons:IfRelationIsActiveTheColumnCannotAlignToQuestionInInvalidPosition}
     	\xFour{\columnVar}{\columnVarPrime}{\qCons}{\qConsPrime} \leq  1 - \xTwo{\tableCell}{\hat{\qCons}}, \\ \textrm{where } \hat{\qCons} \leq \qCons \textrm{ and } \tableCell \in \columnVar
      \end{multline} \\
       %\tushar{These constraints are needed because of some active column/relation vars not being created but are already covered by the prior constraints if we create all the variables. Are we creating such constraints?: If a column is inactive, no relation match connecting to the column can be active, If a relation match is associated with a column, no column connected to it can be active} \daniel{don't get your point! :-/} & 
      %\begin{equation}
      %1
      %\end{equation} \\
      If a row is active, at least a certain number of its cells must be active & 
      \begin{equation}
          \sum_{k} \xOne{\tableCell} \geq \textsc{MinActiveCellsPerRow} \times \xOne{\rowVar}, \forall i, j
      \end{equation} \\
      If row is active, it must have non-choice alignments. & 
      \begin{equation}
          \xOne{\rowVar} \leq \sum_{(n, n') \in \mathcal{L}_{ij}} \xTwo{n}{n}
      \end{equation} \\
      If row is active, it must have non-question alignments & 
      \begin{equation}
          \xOne{\rowVar} \leq \sum_{(n, n') \in \mathcal{K}_{ij}} \xTwo{n}{n}
      \end{equation} \\
      If two rows of a table are active, the corresponding active cell variables across the two rows must match; in other words, the two rows must have identical activity signature & 
      \begin{multline}
          \xOne{\rowVar} + \xOne{\rowVarPrime} + \xOne{\tableCell}  \\ - \xOne{\tableCellPrime} \leq 2, \forall i, j, j', k, k' 
      \end{multline} \\
       If two rows are active, then at least one active column in which they differ (in tokenized form) must also be active; otherwise the two rows would be identical in the proof graph. & 
       \begin{equation}
         \sum_{\tableCell \neq \tableCellSameRow} \xOne{\columnVar} - \xOne{\rowVar} - \xOne{\rowVarPrime} \geq -1 
      \end{equation} \\
      If a table is active and another table is also active, at least one inter-table active variable must be active; & 
      \begin{equation}
          \xOne{\tableVar} + \xOne{\tableVarPrime} + \sum_{j, k, j', k'} \xTwo{\tableCell}{\tableCellPrimePrime}  \geq 1, \forall i, i' 
      \end{equation} \\
      \hline 
    \end{tabular}
    }
%     \caption{The set of all constraints used in our ILP formulation. The set of variables and are defined in Table~\ref{table:ilp-variables}. More intuition about constraints is included in Section 3. The sets used in the definition of the constraints are defined in Table~\ref{table:ilp-sets}.}
% 	\label{table:constraints}
    \caption{The set of all constraints used in our ILP formulation. The set of variables and are defined in Table~\ref{table:ilp-variables}. More intuition about constraints is included in Section 3. The sets used in the definition of the constraints are defined in Table~\ref{table:ilp-sets}.}
	\label{table:constraints}
\end{table*}

\makeatletter% Set distance from top of page to first float
\setlength{\@fptop}{0pt}
\makeatother

\fi 
\end{document}